# The Prompt Engineering Report Distilled: Quick Start Guide for Life Sciences

Valentin Romanov[1#], Steven A Niederer[1,2]

1. National Heart & Lung Institute, Faculty of Medicine, Imperial College London, United Kingdom
2. Turing Research and Innovation Cluster: Digital Twins, The Alan Turing Institute, London, United Kingdom

**# Corresponding author: v.romanov@imperial.ac.uk**

**Abstract:**

Developing effective prompts demands significant cognitive investment to generate reliable, high-quality responses from Large Language Models (LLMs). By deploying case-specific prompt engineering techniques that streamline frequently performed life sciences workflows, researchers could achieve substantial efficiency gains that far exceed the initial time investment required to master these techniques. The Prompt Report published in 2025 outlined 58 different text-based prompt engineering techniques, highlighting the numerous ways prompts could be constructed. To provide actionable guidelines and reduce the friction of navigating these various approaches, we distill this report to focus on six core techniques: zero-shot, few-shot, chain-of-thought, ensembling, self-criticism, and decomposition. We break down the significance of each approach and ground it in use cases relevant to life sciences, from literature summarization and data extraction to editorial tasks. We provide detailed recommendations for how prompts should and shouldn't be structured, addressing common pitfalls including multi-turn conversation degradation, hallucinations, and distinctions between reasoning and non-reasoning models. We examine context window limitations, agentic tools like Claude Code, while analyzing the effectiveness of Deep Research tools across OpenAI, Google, and Anthropic platforms, discussing current limitations. We demonstrate how prompt engineering can augment rather than replace existing established individual practices around data processing and document editing. Our aim is to provide actionable guidance on core prompt engineering principles, and to facilitate the transition from opportunistic prompting to an effective, low-friction systematic practice that contributes to higher quality research.

## 1. Introduction

If the latest flagship models like ChatGPT-5.4-thinking and Claude Opus 4.6 work "out of the box", then why do Large Language Model (LLM) companies provide highly detailed and specific instructions ('GPT-5 Prompting Guide' 2025; Anthropic 2025a) explaining how to prompt their models? The answer to this question is the motivation behind writing this review. The simple answer is that while these models perform exceptionally well for basic tasks, anything requiring precision, such as finding specific references, editing, code, data extraction, model fitting, demands carefully structured prompts. A prompt such as "*improve this paragraph to sound more professional*" is not only ill-defined, lacking specificity, vague; Improve how? What does professional mean? What does 'to sound more' mean in this context? Are there gold standard examples of what professional look like?

Developing systematic questions to improve LLM responses is the essence of prompt engineering, which is the intentional curation and development of a prompt by imbuing it with various conditionals (Schulhoff et al. 2025). By definition, this must mean prompt engineering requires a certain level of cognitive load, reflection and understanding of the problem it is being engineered towards. Most non-AI experts approach prompting opportunistically rather than systematically (Ye et al. 2024). That being said, designing deliberate and effective prompts is hard even for LLM and domain experts (Zamfirescu-Pereira et al. 2023) and is an ongoing area of research in Natural Language Processing (Bommasani et al. 2022) with companies like Anthropic providing an automated generator (Anthropic 2024a) to help reduce friction.

And yet, because an answer to a hard question appears to only be a prompt away, LLM adoption within sciences has skyrocketed. Of the 2534 responses provided to a Danish survey on LLM usage in academia, almost 50% reported using LLMs for "editing a research article for improving readability" and "refining language for research proposals", with ~ 35% using them for writing and editing code, summarizing and analysing literature, or proposing a title, abstract or for generating keywords (Andersen et al. 2025). These usage proportions are higher than they were in 2023 (Van Noorden and Perkel 2023), and as model capabilities improve, it's hard not to envision that more and more work will be offloaded to LLMs in the near future. Within the life sciences, LLMs have been used within the field of cardiovascular disease (Mishra et al. 2024; Sharma et al. 2024), protein-protein interactions (M. Jin et al. 2024; Chang et al. 2025), extracting materials data from research papers (Polak and Morgan 2024), and assisting with chemical synthesis predictions (Zheng, Zhang, Borgs, et al. 2023; Ruan et al. 2024). LLM applications extend to enhancing mathematical reasoning through topic-aware prompt engineering and self-evaluation (Sukherman and Folajimi 2025; Imani et al. 2023), though recent work has shown that chain-of-thought approaches can sometimes reduce performance on certain mathematical tasks (Evstafev 2025). LLMs are being integrated into digital twins (Amad et al. 2025; Sun et al. 2024), for creating gene signalling networks (Tewari et al. 2025) and in drug discovery and development (Y. Zheng et al. 2024).

In this review, we distill the latest advances in prompt engineering techniques for life sciences applications. We provide a detailed examination of core prompting techniques as described within the Prompt Report (Schulhoff et al. 2025), focusing on six categories: zero-shot, few-shot approaches, thought generation, ensembling, self-criticism, and decomposition. Where possible, we provide case studies to illustrate these concepts. We also review published prompts and suggest improvements based on current best practices. Further, we examine studies that have deployed LLMs for tasks ranging from literature summarization and data extraction to protein interaction prediction and synthesis planning. We focus on core prompt engineering principles, building complexity incrementally while introducing relevant nuance, and providing concrete recommendations where possible. By highlighting common pitfalls such as multi-turn conversation degradation, hallucination risks, and the differences between reasoning and non-reasoning models, we aim to guide researchers in selecting and implementing prompting strategies appropriate for their experimental and professional objectives.

We also address emerging frameworks including agentic tools and Deep Research, emphasizing their current limitations alongside potential applications. While predicting the trajectory of prompt engineering remains challenging given rapidly evolving model capabilities, we document current best practices that have remained consistently relevant since the introduction of ChatGPT-3.5 in 2022. We propose that systematically engineered prompts, when properly implemented, may reduce the burden of various research and non-research activities, from literature review and data analysis to providing new perspectives on established workflows in drug discovery, systems biology, and precision medicine. We then conclude with a discussion centred around ethics, bias and data retention policies of the three major AI providers while also highlighting journal policies surrounding AI use. In the end, prompt engineering augments rather than replaces existing pipelines, offering researchers additional tools for interacting with familiar data and processes. Our aim is to provide actionable guidance on core prompt engineering principles, contributing to the transition from opportunistic prompting to an effective, low-friction systematic practice.

## 1.1. Summary of the Prompt Engineering techniques discussed in this review

Table 1 covers the prompt engineering technique, the task it was applied to, the accuracy before and after prompt engineering was applied.

**TABLE 1 Quantitative performance metrics across prompt engineering techniques.**

| Technique | Task | Before | After | Metric | Reference |
|---|---|---|---|---|---|
| ***Zero-shot → Few-shot*** | | | | | |
| Few-shot | Clinical NER (VAERS) | F1 = 0.288 | F1 = 0.676 | F1 score | (Y. Hu et al. 2024) |
| Few-shot (20-shot) | Histopathology classification | 40% | 90% | Accuracy | (Ono et al. 2024) |
| Few-shot + CoT + voting | Medical QA (MedQA) | 58.4% | 72.6% | Accuracy | (Maharjan et al. 2024) |
| Few-shot (structured) | PPI extraction (LLL) | F1 = 30.3% | F1 = 81.6% | F1 score | (Chang et al. 2025) |
| Example diversity | Few-shot in-context | Baseline | +12.9% relative | Performance | (Su et al. 2022) |
| Prompt ordering | Data extraction | Baseline | ±5.5–10.5 pp | Accuracy | (Bhope et al. 2025) |
| Task→Context ordering | Protocol identification | 0% (context→task) | 100% (task→context) | Accuracy (n=10) | *This work, Fig. 8b* |
| ***Chain-of-Thought*** | | | | | |
| Zero-shot CoT | Arithmetic (MultiArith) | 17.7% | 78.7% | Accuracy | (Kojima et al. 2023) |
| Few-shot CoT | Math (GSM8K) | 17.9% | 56.9% | Accuracy | (Jason Wei, Wang, et al. 2022) |
| Few-shot CoT (4-shot) | Arithmetic reasoning | 4% | 90% | Accuracy | (Liang 2025) |
| CoT + reasoning model | Cognitive tasks (o1-preview) | Baseline | −36.3% | Accuracy | (R. Liu et al. 2025) |
| ***Ensembling*** | | | | | |
| Self-consistency | Math (GSM8K) | Baseline CoT | +17.9% | Accuracy gain | (X. Wang et al. 2023) |
| Self-consistency | Strategy QA | Baseline CoT | +6.4% | Accuracy gain | (X. Wang et al. 2023) |
| Composite (Medprompt) | Medical QA (MedQA) | 81.4% (5-shot) | 90.2% | Accuracy | (Nori et al. 2023) |
| Majority vote | Inconsistency detection | 3/9 models fail | 6/9 rescued | Majority verdict (n=5) | *This work, Table 3* |
| ***Self-Criticism*** | | | | | |
| Self-Refine | Code, sentiment, dialogue | Baseline | 5–40% (→20% avg.) | Task performance | (Madaan et al. 2023) |
| Reflexion | Code (HumanEval) | 80.1% | 91.0% | Pass@1 | (Shinn et al. 2023) |
| Self-correct (no tools) | Math (GSM8K) | Baseline | ≤0% (degraded) | Accuracy | (Jie Huang et al. 2024) |
| Self-reflection | Protocol ID (Opus 4.6) | 0% (context→task) | 90% (+reflection) | Accuracy (n=10) | *This work, Fig. 14b* |
| Self-reflection | Protocol ID (Gemini 3.1) | 30% (context→task) | 100% (+reflection) | Accuracy (n=10) | *This work, Fig. 14b* |
| Self-reflection – negative | Protocol ID (ChatGPT-5.4) | 10% (context→task) | 30% (+reflection) | Accuracy (n=10) | *This work, Fig. 14b* |

| Iterative refinement | Pathology report extraction | 1.13% | 0.99% | Major Error Rate | (Hein et al. 2025) |
|---|---|---|---|---|---|
| ***Decomposition*** | | | | | |
| Least-to-Most | Compositionality (SCAN) | 16.2% (CoT) | 99.7% | Accuracy | (Zhou et al. 2023) |
| Least-to-Most | Numerical reasoning (DROP) | 74.8% (CoT) | 82.5% | Accuracy | (Zhou et al. 2023) |
| Plan-and-Solve (PS+) | Arithmetic (6 datasets avg.) | 70.4% (ZS-CoT) | 76.7% | Accuracy | (L. Wang et al. 2023) |
| DecomP | Math (MultiArith) | 78% (CoT) | 95% | Accuracy | (Khot et al. 2023) |
| ***Personas*** | | | | | |
| Role-play prompting | Reasoning (AQuA) | 53.5% (zero-shot) | 63.8% | Accuracy | (Kong et al. 2024) |
| Expert persona | Factual QA (GPQA Diamond) | ~30% (no persona) | ~30% (physics expert) | Accuracy | (Basil et al. 2025) |
| Expert persona | MMLU (2,410 questions) | Control | No significant Δ | Accuracy | (M. Zheng et al. 2024) |
| ***Multi-turn*** | | | | | |
| Multi-turn degradation | Combined (aptitude+reliability) | 90% (single-turn) | 65% | Performance | (Laban et al. 2025) |

Negative and null results are included to highlight techniques and contexts where prompt engineering provides no benefit or degrades performance.

## 2. Prompting and Prompt Engineering

Our discussion on prompting will focus specifically on what it means to prompt and what it means to prompt engineer. Programmatically automating and developing strategies for iterating on and improving prompts can be complex and is not how most academics engage with LLMs and as such, is not covered here. Creating thoughtful, fully specified prompts can take significant time and can be frustrating. The goal of this review is to provide comprehensive demonstrations of how to engineer prompts, with as little friction as possible, to allow academics to move quickly and confidently.

Core prompting techniques can be broken down into: 1) Text based Techniques, 2) Multilingual Techniques and 3) Multimodal Techniques (audio, video etc.) (Schulhoff et al. 2025). The core focus of this review will be on text-based techniques. We will mention considerations for getting reliable results from multilingual prompting and plan to release a separate report on multimodal techniques in the future. In the Prompt Report, the authors identified 58 different prompting techniques. These techniques can be broken into 6 major categories, covering the various unique ways to approach constructing prompts. In this section, we will focus on these 6 categories and provide a flavour of each prompting technique with research-specific use cases.

### 2.1. Zero-shot

There are several ways to think about prompting engineering. As a thought experiment, we can directly map two different but common approaches to LLM usage patterns. Exploratory searches, such as reading an initial PubMed abstract to understand a topic, align with zero-shot prompting where one can enquire and rapidly obtain high-level understanding of the concept. On the other hand, using structured, well-specified prompt engineering approaches like few-shot, ensembling and decomposition require significantly more effort to engineer and are more akin to a systematic review.

Zero-shot refers to the case where the user simply provides a prompt (a text based instruction) to the LLM without defining what the output should look like. In this case, the assumption is that the model will have seen similar data in its dataset during training. The best way to structure a zero-shot prompt is to provide the request or question together with rules (Wei, Bosma, et al. 2022) or to specify what it can and can't do. While there are many zero-shot strategies one can deploy, we refer the user back to the original Prompt Report. Here we will discuss the most frequently used techniques deployed within academic work where LLMs are utilized for summarizing articles, extracting information, or editing manuscripts.

Not all prompts are equal, especially within academia. We provide 5 examples of fairly simple prompts: 2 prompts demonstrate where a simple request will give the user satisfactory outcomes and another 3 prompts that showcase how language can drive poor outcomes (Figure 1). The top row of Figure 1 shows two domains where LLMs perform well, 1) providing information regarding domains where they have seen a lot of data during pre-training and 2) even if they lack the training data, they can use tool calling (internet search) to verify output. Prompts that are not well specified may force the LLM to guess user intent, for example, asking the model to summarize an article is bound to disappoint if the model doesn't understand at which level to summarize. Prompts that lack specifics (target audience, examples of what "good" looks like etc.) or are too broad (do not tell the model what to focus on) force the LLM to act more and more as a "useful assistant" and because LLMs are not encouraged to respond with "I don't know" what they usually respond with is something that is plausible (hallucinations) (Kadavath et al. 2022). Further, learning from LLMs, while effective in some domains, is less effective within domains with little training data, for example, a PhD student trying to learn a nuanced concept published earlier on in the year. LLMs are unreliable within work environments that push the frontier of knowledge.

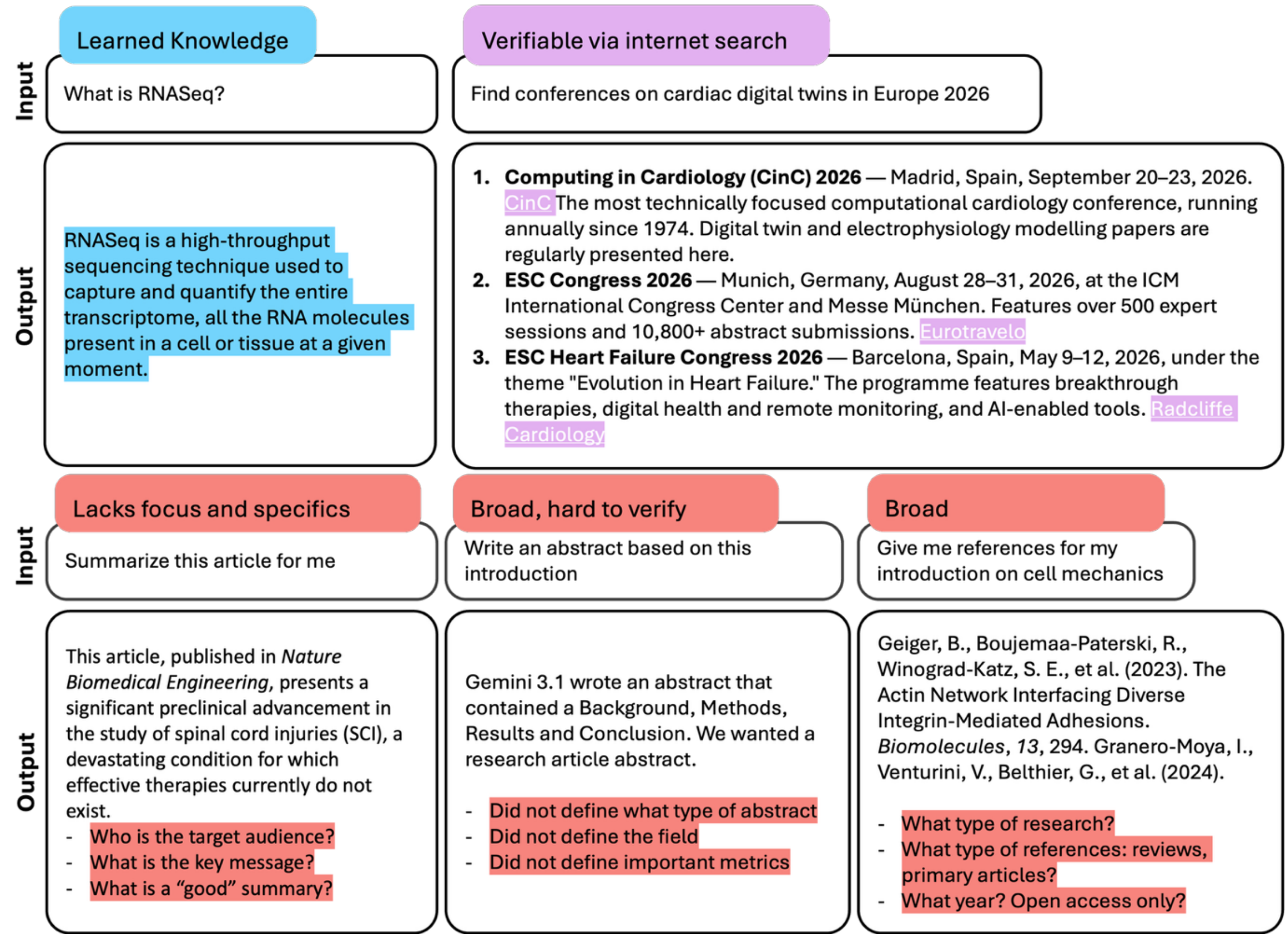

**Figure 1. Examples of Zero-Shot prompting**. Some prompts do not need to be overly complicated, as long as the domain has enough training data or can be verified, prompts do not need to be overly prescriptive. However, asking the model to "summarize" or express a form of "taste", express an opinion, or a judgment or guess a preference, will typically result in sub-optimal output due simply to the fact that the model is then forced to respond based on what it understand of the topic, at that point, with the reasoning traces present for that conversation.

Generating article summaries is a popular demonstration of zero-shot prompting and significantly useful for academic work. Well-crafted summaries can help researchers narrow down the potential list of articles that need to be read to gain the necessary coverage of a topic and to reduce fatigue. In a recent study, Hanson et al. (2024) showed that total articles indexed in Scopus and Web of Science have grown exponentially in recent years; in 2022 the article total was approximately ~47% higher than in 2016, which has outpaced the limited growth in the number of practising scientists (Hanson et al. 2024). As such, LLM summaries that capture the coverage, context and nuance of research articles can be extremely useful.

Can we trust LLM summaries of academic research? In short, current research does not support this. While LLMs have been extensively used for summarizing research articles, current research suggests that these summaries lack nuance. Recently, Peters and Chin-Yee (2025) tested a number of frontier LLMs (Sonnet 3.7, ChatGPT-4o, DeepSeek, amongst other models) in their ability to summarize articles across multidisciplinary sciences and medical journals, finding that LLM-generated summaries were 5 times more likely to contain overgeneralizations compared to human-authored summaries (Peters and Chin-Yee 2025). Similarly, LLM generated abstracts, while more readable than the original, lacked quality as measured by the CONSTOR-A checklist (Hwang et al. 2024).

A goal of this review is to bring together the latest ideas on prompt engineering and provide recommendations where possible to enhance past work. In Figure 2, we have taken the original prompts given by the authors and included suggestions that may improve the adherence, specificity and output of the LLMs. Our analysis reveals several areas for improvement across all prompt types. First, the "simple prompt" lacks field specificity and examples. Providing the LLM with multiple examples of high-quality human-generated summaries in the relevant field, along with examples of poor summaries, helps guide LLM output. Second, while the systematic prompt introduces step-by-step reasoning, it would benefit from concrete examples of effective thought processes for information retrieval, allowing the model to better understand what is expected. Third, the accuracy prompt's use of negative instructions ("do not introduce any inaccuracies") can be an issue as LLMs cannot reliably self-assess factual accuracy. Instead, specific guidelines about what to include or how to verify information would be more effective. Finally, the full text prompt highlights critical technical constraints that require careful navigation. Processing entire research articles demands careful monitoring of token consumption; one effective strategy is to provide the LLM with one well-specified prompt and a single article per conversation to maintain focus and accuracy. A constant theme throughout this review is that creating well-specified prompts is hard, requiring not just clear instructions but also concrete examples, field-specific context, and awareness of technical limitations.

| Original | Suggested Improvements |
| --- | --- |
| **Simple prompt:**<br>"I'm going to provide an abstract of a research article. Please provide a summary of the main findings mentioned in the abstract." | **Simple prompt:**<br>+ Specify which field<br>+ Provide multiple examples of what good human-generated summaries in a similar field look like<br>+ Provide multiple examples of what bad summaries look like |
| **Systematic prompt:** "I'm going to provide an abstract of a research article. Please provide a summary of the main findings mentioned in the abstract. Take a deep breath and work on this problem step-by-step." | **Systematic prompt:**<br>+ All the above<br>+ Provide multiple examples of good step-by-step thought processes for retrieving the necessary information |
| **Accuracy prompt:** "Generate a comprehensive and information-dense summary of the following paper abstract. Include key points, research questions, and conclusions of each paper, and structure the information in a logical and easy-to-understand format. Stay faithful to the source material and do not introduce any inaccuracies when summarizing." | **Accuracy prompt:**<br>+ All the above<br>+ Avoid the use of **"Do not"** without specifying what to specifically to avoid, in this case the LLM doesn't know its not being factual |
| **Full text prompt:** "I'm going to upload a research article. Please provide a detailed summary of the findings mentioned in the article. Please also provide a title for your summary that captures the main finding." | **Full text prompt:**<br>+ All the above<br>+ Provide 1 well specified prompt, and 1 article per conversation<br>+ Ensure the LLM has the context window to process the request |

**Figure 2. A zero-shot prompt case study**. Original prompts from Peters and Chin-Yee (2025) (Peters and Chin-Yee 2025) are shown on the left, with suggested improvements on the right. Each improvement (+) represents a specific enhancement that may improve outcomes: adding domain specificity, incorporating quality exemplars, clarifying instructions to avoid ambiguity, and addressing technical constraints.

## 2.2. On Context Window and Token Consumption

The context window of an LLM refers to the number of tokens an LLM can attend to (remember) at once. When passing a text phrase into an LLM, the text is encoded into a sequence of token ids, mapping to substrings (Rajaraman et al. 2025). One token roughly represents 0.5 words (based on Gemini tokenization ratio via AI Studio). Exceeding the length of the context window typically results in hallucinations, and severe performance degradation (An et al. 2024). Further, LLMs are typically not able to utilize the entire context window, remembering information in the beginning and at the end of the context window, but overlooking information in the middle (H. Jin et al. 2024). Based on publicly available information, the free version of ChatGPT has a context window of 16k tokens, Gemini has a context window of 32k tokens while Claude has 200k tokens. A typical research article is roughly 4k tokens. In Figure 3 we illustrate the relative difference in context size between the free offerings for 3 different LLM providers, where the sphere volumes are proportional to their respective context windows. This visualization demonstrates that while ChatGPT's free tier can accommodate approximately 4 research articles, Gemini can process 8 articles, and Claude's substantially larger context window can handle approximately 50 articles simultaneously. This dramatic difference in capacity has deep implications for conducting reliable, cohesive, and comprehensive literature reviews, while working with multiple documents and document types.

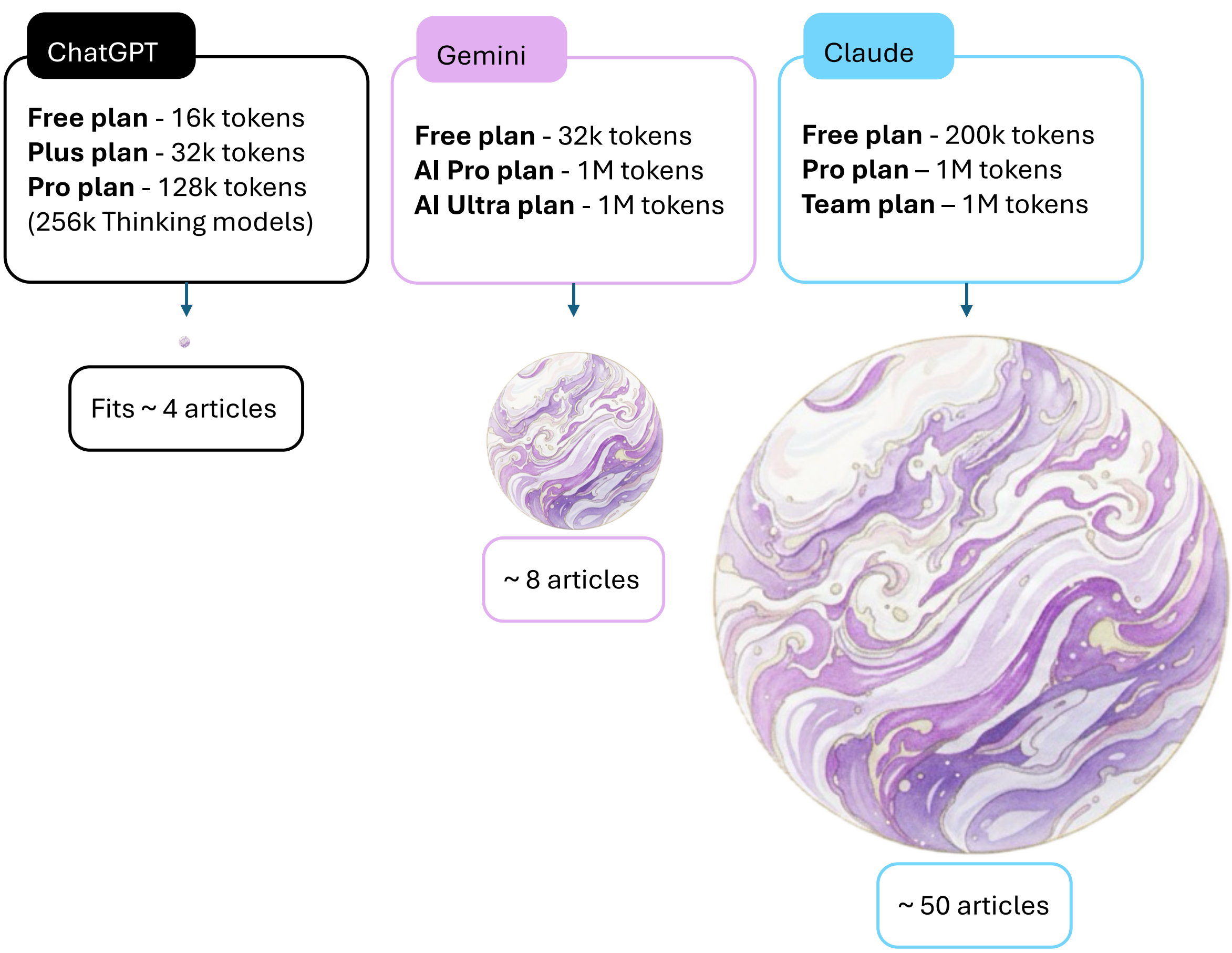

**Figure 3. Comparative visualization of context window capacities across free-tier LLM offerings.** Token limits for ChatGPT (16k tokens), Gemini (32k tokens), and Claude (200k tokens) are shown with their respective subscription tiers, alongside a volumetric representation where sphere sizes are proportional to context window capacity. The visualization translates these token limits into practical research contexts, demonstrating that the free versions can accommodate approximately 4 articles (ChatGPT), 8 articles (Gemini), and 50 articles (Claude) based on an average research article length of 4k tokens. Claude's free tier offers 12.5x the capacity of ChatGPT and 6.25× that of Gemini.

### 2.3. Personas

Assigning a persona to an LLM is one of the most popular use cases of this technology, for example, “Take on the role of a clinician” or “Take on the role of a literature review specialist”. Persona assignment is used frequently within and outside the realm of academia, Character.ai, a website hosting thousands of AI personalities has around 20+ million monthly visitors. The most prominent use of persona assignment can be found in system prompts provided to LLMs by all major providers. System prompts describe how the chatbot will behave and interact with the user. For example, the system prompt for Claude Opus 4.6 from Anthropic outlines how the model should help the user with improving prompts, how to handle difficult people, even how to provide emotional support (Anthropic 2024b).

Personas represent abstractions that emerge from statistical patterns in the training data. These abstractions often amplify latent biases and stereotypes present in the source material. Figure 4 illustrates this phenomenon through Claude Opus 4.6's interpretation of a "clinician" persona. When presented with an identical prompt across three independent conversations, the model generated responses that, while thematically consistent (empathetic, evidence-based, direct, precise, trustworthy etc.), exhibited notable variation in the specific attributes emphasized, ranging from "calm" and "patient" in Response 1 to "authoritative" in Response 2. Our initial prompt lacks specificity such as clinical specialty, practice setting, geographic location, or level of seniority. The model's responses thus reflect not a singular, well-defined professional identity, but rather a composite representation drawn from diverse clinical archetypes within its training data.

**Original**

**Prompt**
"I want to create a useful chat bot. It will take on the persona of a clinician, in less than 10 keywords, what is the persona of a clinician?"

| | Response 1 | Response 2 | Response 3 |
|---|---|---|---|
| **Empathetic** | ✓ | ✓ | ✓ |
| **Evidence-based** | ✓ | ✓ | ✓ |
| **Direct** | ✓ | ✓ | ✓ |
| **Non-judgmental** | ✓ | ✓ | ✓ |
| **Precise** | ✓ | ✓ | ✓ |
| **Trustworthy** | ✓ | ✓ | ✓ |
| **Methodical / Systematic*** | ✓ | ✓ | ✓ |
| **Calm** | ✓ | — | ✓ |
| **Authoritative** | — | ✓ | ✓ |
| **Patient** | ✓ | — | — |
| **Cautious** | — | — | ✓ |

**Figure 4. Variability in LLM persona interpretation across independent conversations.** The same prompt requesting Claude Opus 4.6 to describe "the persona of a clinician" in under 10 keywords was submitted in three separate conversation threads. Despite the identical input, the model produced three distinct sets of attributes, demonstrating both semantic overlap (e.g., "empathetic," "direct," "precise" appearing across responses) and notable divergence in emphasized characteristics (e.g., "calm" vs. "patient" vs. "cautious"). Underspecified persona prompts can yield inconsistent and unreliable outputs.

Figure 5 showcases two role-playing prompts for chemistry applications. The first prompt (Figure 5, left) clearly defines to the LLM its function and rules in order for the LLM to function as a chemistry assistant extracting synthesis parameters into JSON format (Silva et al. 2024). By chaining multiple well-defined LLM instances, it is possible to assemble a team of LLMs to work towards a solution, together. Prompt 2 outlines how this can be deployed (Figure 5, right) where the authors give the LLM the persona of Bohr, a literature review specialist within a seven-member AI team optimizing aluminium-based metal-organic frameworks crystallinity (Zheng, Zhang, Nguyen, et al. 2023). Each team member, from Project Manager Atlas to the Robot Technician, has distinct responsibilities, enabling specialized task distribution.

The two prompting strategies along with persona assignment are different. Multi-agent systems like Bohr's team excel when problems require diverse expertise and iterative refinement, such as optimizing synthesis conditions through Bayesian optimization. In contrast, the standalone assistant with explicit output examples proves more effective for standardized data extraction from literature. Both approaches include examples of expected outputs. Examples help to reduce ambiguity and improve output consistency.

Throughout this review, we employ specific terminology with precise operational definitions. Large Language Models (LLMs) refer to the transformer-based neural architectures trained on extensive text corpora (ChatGPT, Claude, Gemini). Agents or agentic systems denote LLMs augmented with tool-use capabilities, memory systems, and autonomous task execution frameworks such as Deep Research, Claude Code (Anthropic), and similar implementations.

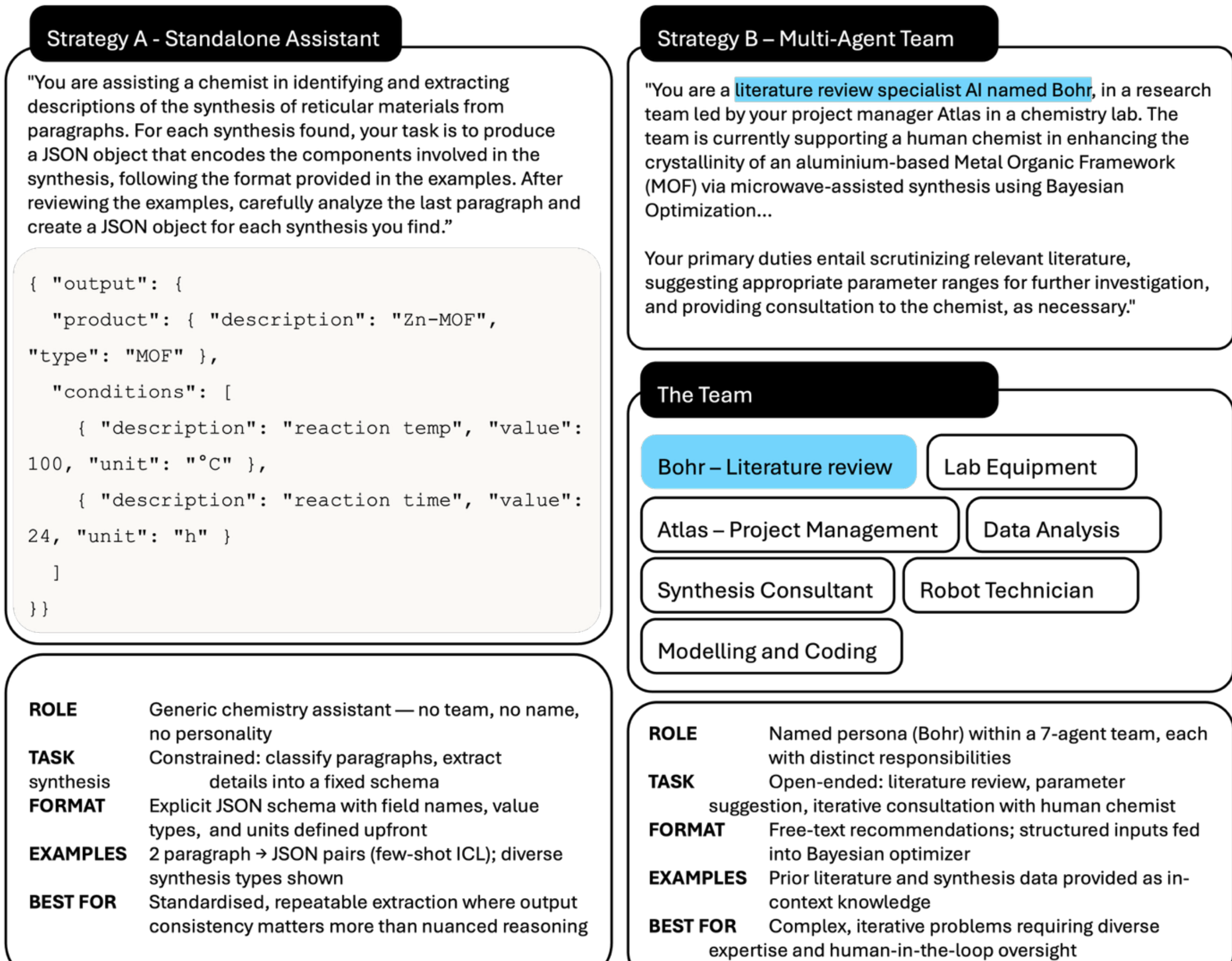

**Figure 5. Examples of a persona and an assistant prompt from literature.** Left: A chemistry assistant prompt for extracting information regarding the synthesis of reticular materials from information provided by the user. The assistant prompt has two parts, a description of the task and an example of the output (Silva et al. 2024). Right: An example of a literature review specialist named Bohr, as part of a larger team of AI agents, each with their own persona, including Project Manager, Analytical Assistant, Chemical Synthesis Consultant, Modeling and Coding Specialist, Robot Technician, Lab Equipment Designer, and the Data Analysis Assistant Specialist (Zheng, Zhang, Nguyen, et al. 2023). Right: Adapted with permission under the Creative Commons Attribution 4.0 International License (CC BY 4.0).

Is there a difference between asking the LLM to take on a persona vs acting as an assistant? The scientific literature is yet to settle on this point. Evaluations are made harder because some articles refer to personas as assistants. The current state of the field suggests there may be some advantages to using expert personas that work within their domain, however the effect is small and performance is typically unreliable (Araujo and Roth 2025). While some suggest there is room for sculpting distinct AI personalities, as it stands, assigning personas to an LLM on objective tasks such as mathematics and physics, typically results in unpredictable behaviour (M. Zheng et al. 2024). Li et al. (2025) suggested the need for "rigorous science" around persona generation finding that personas representing the general population for predicting voting patterns were shown to be significantly biased (A. Li et al. 2025).

One interesting emerging application of personas is in population and disease modelling. Although LLMs like Gemini 3.1 Pro, Claude Opus 4.6 and ChatGPT-5.4 can generate informative prior distributions for explaining cardiovascular health outcomes, the reliability of this approach is inconclusive. Although these models can generally identify the correct directional associations (e.g., whether a factor increases or decreases disease risk), they struggle with calibrating the magnitude and confidence intervals of these associations (Riegler et al. 2025). This contrasts with Capstick et al. (2025), who successfully leveraged LLMs to generate prior distributions that significantly accelerated predictive model development, reducing the required manual labelling time by more than six months (Capstick et al. 2025).

### 2.4. Stylistic Guidance

The LLM can be guided via prompts to edit sentences, change tone, and stylistic direction of the text (Lu et al. 2023). LLM adoption within the scientific community for helping write, and translate academic text has been astronomical (Van Veen et al. 2024). Using LLMs "to help write research articles" was the 4$^{th}$ most popular use case, accounting for ~28% of use by researchers, as reported in a Nature survey (Van Noorden and Perkel 2023). An analysis of biomedical research literature discovered that around 13.5% of abstracts (~200,000 papers) published in PubMed in 2024 used an LLM. Usage varied greatly (lower bound between 5% to more than 40%) between research fields, affiliated countries and journals (Kobak et al. 2025). In 2025, the majority of researchers seem to employ LLMs primarily as editing tools rather than for complete article generation, which may enhance the clarity and accessibility of their research (DeHaan et al. 2025).

Whether using LLMs for these tasks is good for science in general is unclear. From biases (Bender et al. 2021), to fabricating references (Walters and Wilder 2023), there are significant limitations that the user should be aware of. However, as model training scales and new algorithmic methods are added to pre-training and post-training, LLMs are becoming substantially more capable and reliable. While LLMs can still hallucinate references using state-of-the-art models, like Gemini 3.1 Pro and Claude Opus 4.6, it is becoming rarer, and the quality of references that these models suggest, has significantly improved.

While adoption is increasing, there is little guidance or best practices on how to utilize LLMs for editing scientific writing. We provide a structured framework below for leveraging LLMs as editing tools, demonstrating how specific prompt engineering techniques can guide models to perform targeted improvements in grammar, style, and word choice while preserving technical terminology and authorial voice (Figure 6). By requiring the model to quote original text, we check for hallucinations. This is an effective approach when working with either text or data.

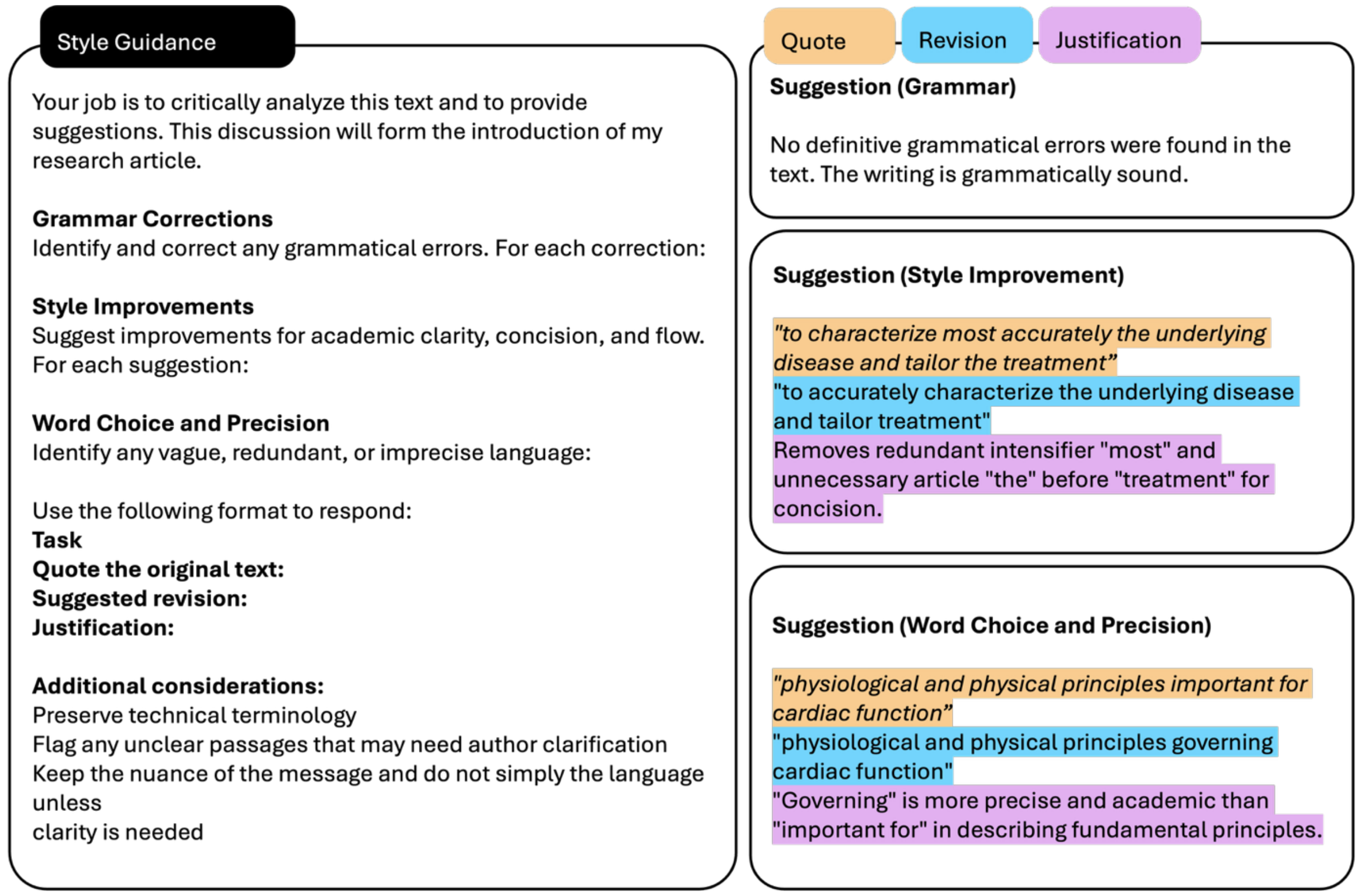

**Figure 6. Recommended approach for leveraging LLMs for editing text**. The framework illustrates prompt engineering strategies for different editing tasks (grammar corrections, style improvements, and word choice refinement), with examples showing original text (orange), suggested revisions (blue), and justifications (purple) to maintain transparency during the editing process.

### Recommendations for engineering effective zero-shot prompts:

- Be explicit. Give the LLM context as to what it is working with, its significance and where it will be used. Here, we pulled part of the introduction from one of our research articles (Niederer et al. 2019). By letting the LLM know it's part of an introduction, there is less chance of non-compliance as it doesn't need to figure out where this may belong within the broader conversation
- Ask the LLM to look over your text and to make suggestions. Its justification for why it thinks this passage requires modification is an important window into understanding how it arrived at its conclusion
- Always ask the LLM to provide "ground truth". Here, we ask the LLM to quote directly from the article. This reduces the chance of hallucinations and is immediately verifiable by the user
- Ideally, provide examples of what good Grammar, Style and Word choice looks like. Provide examples of past introductions that had the qualities you would like to imbue your text with

- Provide additional considerations. An important consideration for scientific editing is to ensure the LLM does not simplify the language, that it preserves field-specific terminology. While stating this in the prompt helps, providing examples of field specific terminology is significantly more useful
- Beyond editing, LLMs can help with overcoming writing barriers (Aydin et al. 2023), and refine and proof-read communications from non-English speaking researchers (Smith et al. 2024).

### 3. Few-shot

In few-shot prompting, we provide the LLM with a small number of examples (typically 2-10) that demonstrate the desired input-output structure for a given task. This approach leverages the model's in-context learning capabilities, allowing it to infer task patterns without updating its weights. The number of demonstrations significantly impacts performance where even 2-3 examples can substantially improve task understanding compared to zero-shot approaches. Recent work has shown that scaling to many-shot regimes (hundreds to thousands of examples) can further improve output quality.

Specifically, providing extensive demonstrations has been shown to enhance generalization across diverse tasks, enable solutions to more complex problems requiring multi-step reasoning, and potentially mitigate certain biases acquired during pre-training (Agarwal et al. 2024). Moreover, this many-shot approach can reduce reliance on computationally expensive fine-tuning procedures, though at the cost of increased context length and potential exposure to adversarial exploits as demonstrated in recent jailbreaking studies. Prompt engineering follows a structured format where the task specification is accompanied by representative input-output pairs that implicitly state task requirements (Figure 7).

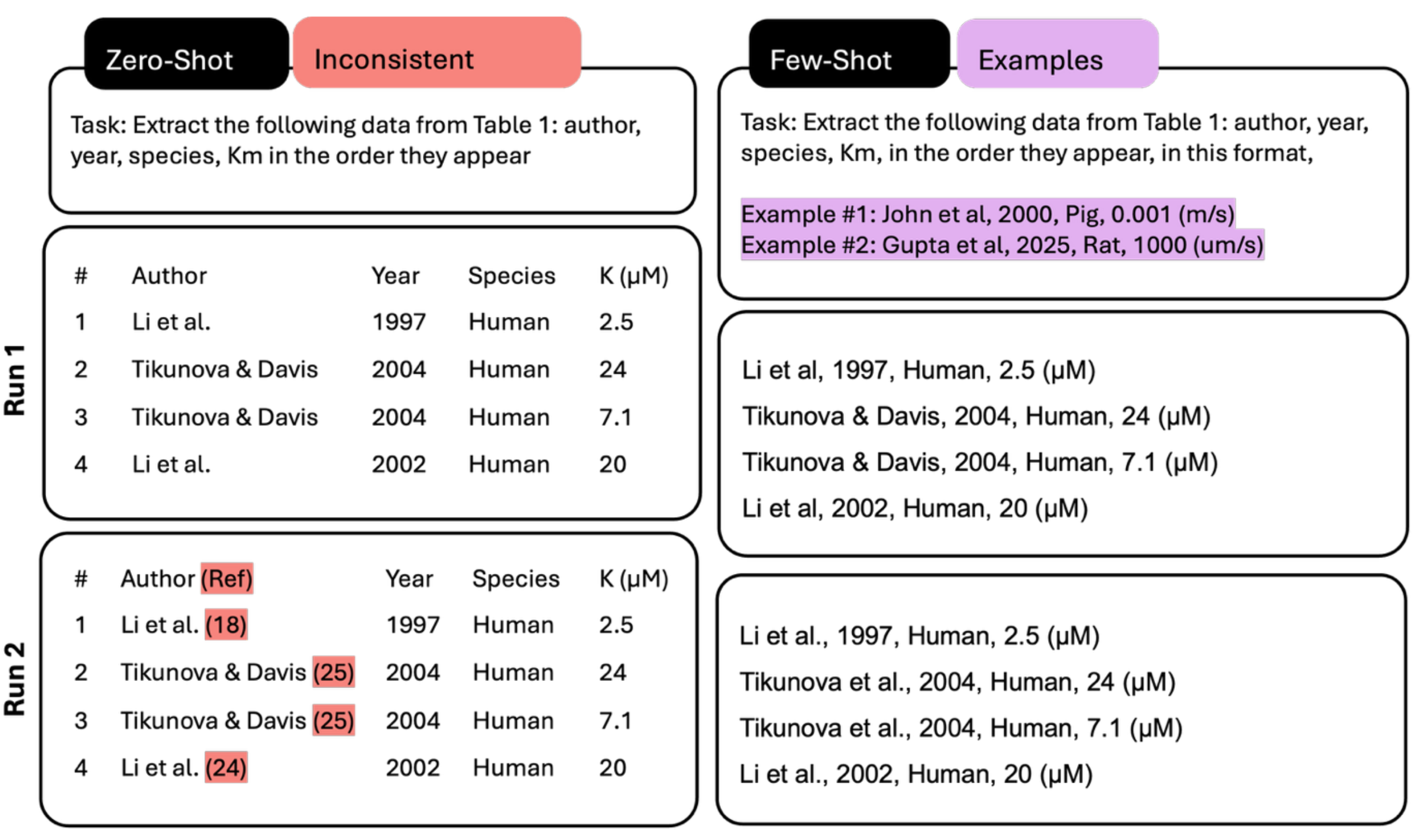

**Figure 7. Comparison of zero-shot and few-shot prompting strategies for structured data extraction.** In zero-shot prompting (left), only the task instruction is provided without examples. The LLM decides how to respond, here for example, Gemini 3.1 Pro decided to add numbered references in run 2 from the table it extracted this data from (S. A. Niederer et al. 2006). In few-shot prompting (right), the same task is augmented with demonstrative examples showing the expected output format and data transformation. Both runs are consistent with the examples. Note the examples illustrate both formatting consistency (author citation style, year, species) and unit handling variability (m/s vs μm/s) that the model must learn to generalize from the provided context.

The quality, and repeatability of the output depends on several factors including:

1) The number of examples provided to the LLM. Typically, the more examples, the better (Brown et al. 2020), while ensuring to monitor overall token count. Agarwal et al. (2024) found that using many examples (>1000 (~85k tokens) in some cases) led to dramatic improvements in summarization, mathematical problem solving, algorithmic reasoning among many other tasks (Agarwal et al. 2024).

2) LLMs are sensitive to the order of information in prompts. Recent evidence demonstrates that the sequential arrangement of in-context examples can substantially impact model performance, with accuracy variations of 5.5-10.5 percentage points depending on example ordering alone (Bhope et al. 2025). This positional bias appears particularly pronounced in tasks requiring structured information extraction from scientific literature. Figure 8 illustrates an experiment to showcase the importance of where to include the task in relation to the context needed to carry out the task.

Figure 8a demonstrates the order in which the task and context was given to the model and the subsequent output by the model. Inspiration for this experiment came from Google's Gemini Prompt Engineering Guide on Vertex, specifically recommending "When dealing with sufficiently complex requests, the model may drop negative constraints (specific instructions on what not to do) or formatting or quantitative constraints (instructions like word counts) if they appear too early in the prompt. To mitigate this, place your core request and most critical restrictions as the final line of your instruction. In particular, negative constraints should be placed at the end of the instruction. A well-structured prompt might look like this: [context] + [main instructions] + [constraints]" (Google Cloud Doc. 2026).

We used a large experimental section from Bax et al. 2025 to provide a part of the context needed to carry out the task and other information that was not directly related to the task (Bax et al. 2025). The task was to answer "What protocol was used for creating iVSMCs suitable for tissue engineering?". We consistently found that placing the task before the context dramatically improved model recall. For each frontier LLM tested (Figure 8b) placing the task first allowed the LLM to correctly interpret the question, "What protocol was used for creating iVSMCs". Placing the task after the context consistently resulted in the model focusing on answering "What protocol was used for tissue engineering with iVSMCs" (Figure 8b).

Further, we did not find any difference between placing the constraints before or after the main context (Figure 8c). In all 10 runs, ChatGPT-5.4 consistently followed the rule to "not explain" and to "simply list the reagents". Claude Opus 4.6 was the most verbose followed by Gemini 3.1 Pro. Opus consistently added commentary or periphery information not related to the task of extracting data with this occurring 80% of the time when the task was placed before the context and 50% of the time when the task was placed after the context. Our results are contrary to the best practices outlined by Google for Gemini for this particular task. Ordering matters and we encourage the user to experiment with this fact.

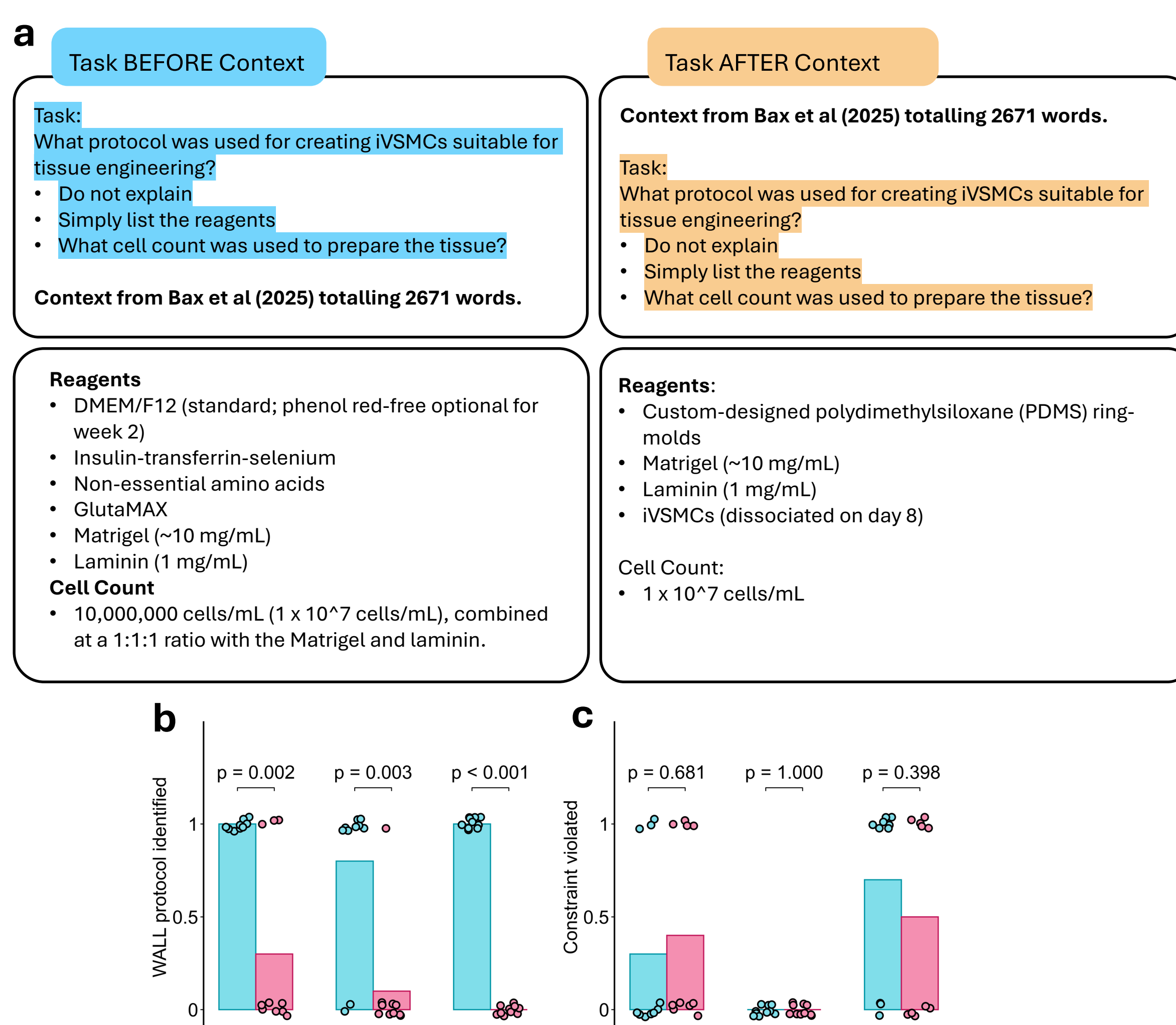

**Figure 8. Effect of task placement order on LLM question interpretation and constraint adherence.** (a) Schematic of the two prompt orderings tested. The task ("What protocol was used for creating iVSMCs suitable for tissue engineering?"), output constraints, and sub-questions were placed either before (left) or after (right) a 2,671-word experimental section from Bax et al. (2025). Representative model outputs are shown below each condition. (b) Proportion of runs in which each model correctly identified the WALL protocol (iVSMC creation) rather than the tissue engineering protocol. Cyan, task placed before context; pink, task placed after context; n = 10 independent runs per condition per model. P values from two-sided Mann-Whitney U test. (c) Proportion of runs in which output constraints ("do not explain"; "simply list the reagents") were violated when constraints were placed before versus after the main context. Conditions and statistical tests as in (b). Individual data points are shown alongside bars representing group proportions.

3) Effective LLM-based extraction requires diverse in-context examples that reflect real-world data heterogeneity. Homogeneous example sets with perfectly formatted, complete entries result in brittle prompts that fail when encountering messy, poorly structured texts. Figure 9 contrasts these approaches: uniform examples versus strategically varied ones that include missing fields, alternative units (48h vs. 48 hours), concentration ranges, and embedded information within complex sentences. This diversification may significantly improve output quality and prompt adherence, as models trained on varied examples develop more robust pattern recognition for incomplete or differently formatted data.

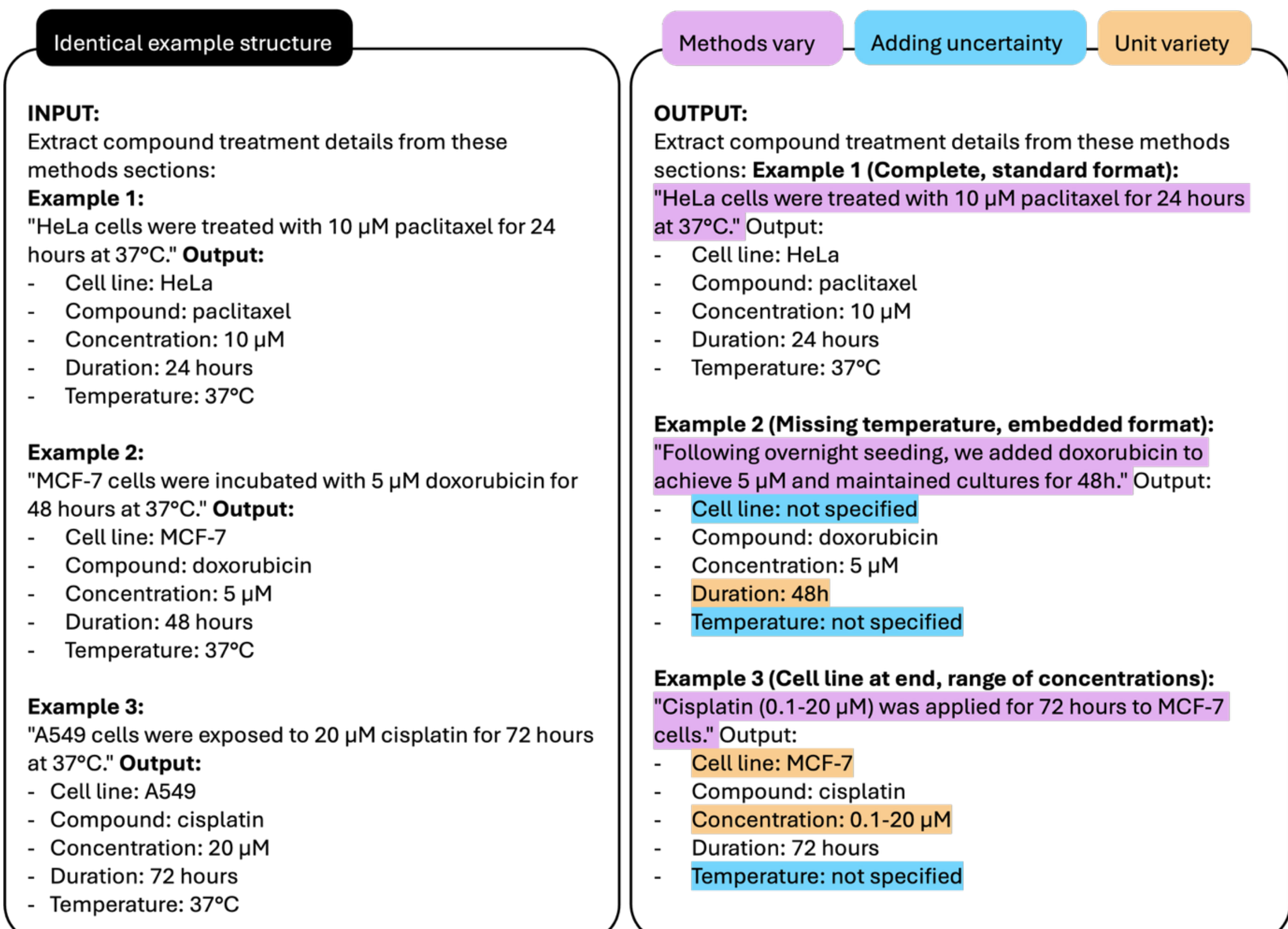

**Figure 9. Example diversity improves extraction robustness.** Left: Homogeneous prompt with three identically structured examples containing complete experimental parameters. Right: Diversified approach incorporating realistic variations with missing data fields (cell line, temperature), embedded information in narrative text, concentration ranges (0.1-20 μM), and a variety of units. Colour coding indicates variation types: purple (structural differences), cyan (missing/uncertain data), orange (unit variety).

### **Recommendations**:

- Avoid using the same patterns, in the example on the left we used: [Cell line] + were + [verb] + with + [concentration] + [compound] + for + [duration] + at + [temperature]. With many examples that follow this order, we bias the model to look for this order and if it can't find it, it may have a harder time extracting information

- Avoid using the same constants, for example using 37 C for each example may cause the model to hallucinate to extract 37 C even if its not in the article or may force the model to return “not specified” because it couldn’t find it
- Provide a variety of representations of how cell lines may be shown within the article
- Include a variety of concentrations and units (uM, mM, mmol/L, mg/L etc.). Avoid asking the LLM to convert to specific units unless multiple examples showcase how to do this well.

4) Domain-specific training examples are surprisingly unnecessary for effective data extraction. State-of-the-art LLMs can successfully transfer extraction patterns across disparate scientific fields, relying on structural templates rather than domain-matched examples. Figure 10 demonstrates this cross-domain transferability where a prompt designed for cell biology experiments, with examples featuring neurons, siRNA transfection, and calcium imaging, successfully extracts analogous parameters from material science (thin film annealing) and analytical chemistry (chromatography). The model maps conceptual categories (sample→material, treatment→conditions, measurement→analytical method, timepoints→duration) across domains without requiring field-specific training.

Source Domain – Cell Biology

Semantic Fields

**Prompt Instructions**

**Extract the following data from this methods section. Match the verbosity of the examples:**

1. Sample/Material : HEK Cells
2. Treatment/Conditions: Triton X-100
3. Measurement Method: Nikon Eclipse TE2000-U
4. Time Points: 24 hrs

**In-Context Examples (Cell Biology Focus)**

**EXAMPLE 1**

"Primary neurons were isolated and cultured for 7 days, then exposed to 100µM glutamate. Calcium imaging was performed at 0, 5, 15, 30, and 60 minutes.”

1. primary neurons (number not specified)
2. 100µM glutamate
3. calcium imaging
4. 0min, 5min, 15min, 30min, 60min

**EXAMPLE 2**

"MCF-7 cells were transfected with siRNA targeting BRCA1. RNA was extracted at 12, 24, 48, and 72 hours for qPCR analysis."

1. MCF-7 cells
2. BRCA1 siRNA transfection
3. qPCR analysis
4. 12h, 24h, 48h, 72h

**Material Science** – Albert et al. 2025

“Then, the thermoelectric thin films were annealed at temperature range from 100°C to 250°C for 30 and 60 min under...(please see reference in figure caption)”

**Semantic Field Mapping:**

1. Bi2Te3 thin films (0.4 µm thick) on SiO2/Si substrates
2. Thermal annealing at 100-250°C under N2 atmosphere
3. XRD, SEM, Seebeck coefficient, electrical conductivity measurements
4. 30 min, 60 min

**Analytical Chemistry** – Molenaar et al. 2022

“The gradient-NPLC $^{1}$D was thermostated at 23 °C, and a flowrate of 40 µL·min$^{-1}$ was applied with a gradient of 30% (v/v) dichloromethane in hexane (mobile phase A) to 5% (v/v)...(please see reference in figure caption)”

**Semantic Field Mapping:**

1. Not specified in this section
2. Gradient from 30% dichloromethane in hexane to 5% THF in dichloromethane
3. NPLC-SEC with DAD and ESI-TOF-MS
4. 0min, 37.5min, 45min, 46min, 65min (gradient program)

**Figure 10. Cross-domain transferability of structured extraction prompts.** Left: Cell biology-specific prompt with examples from neuron culture and transfection experiments. Right: Successful application to unrelated domains such as material science (Albert et al. 2025) ($Bi_2Te_3$ thin film processing) and analytical chemistry (Molenaar et al. 2022) (gradient NPLC-MS). Despite zero domain-relevant examples, the model

correctly maps structural categories across fields, extracting temperature ranges, time points, and measurement methods.

It is generally good practice to ensure that the examples are similar to what the model may expect to encounter in the data it is given (Liu et al. 2021) but contain enough diversity in examples to account for many different ways that information can be presented, summarized and articulated (Wang et al. 2024). In fact, with just 18 examples, Su et al. (2022) showed 12.9% relative gain in performance when selecting examples that were both representative of the task and diverse in coverage within context for the task at hand (Su et al. 2022).

This contrasts markedly with scenarios involving poorly defined information retrieval. When prompt examples contain factual errors or methodological inconsistencies, model accuracy deteriorates significantly with studies demonstrating performance degradation of up to 80% when examples contain incorrect information (Yoo et al. 2022). As an example, a researcher has collected several studies where a specific experiment was analysed using an appropriately justified statistical test. If the statistical tests applied to the experiments are inappropriate, and these examples are provided to the LLM, the model is likely to learn these mistakes and in turn provide incorrect information or hallucinate incorrect answers.

**Recommendations:**

- When extracting information from research articles or other structured texts, focus on developing prompts with clear, consistent formatting. Provide multiple representative examples that demonstrate the exact output structure you require.
- The impact of incorrect examples varies dramatically with model scale. Smaller models (such as PaLM-8B or GPT-3 ada/babbage) rely heavily on their pre-trained knowledge and will largely ignore examples that contradict their semantic priors. This means incorrect examples may have minimal impact on their outputs. Conversely, larger state-of-the-art models (PaLM-540B, GPT-4) actively learn from the patterns in your examples and will override their pre-training to follow the examples you provide, even when those examples are incorrect (Wei et al. 2023).
- When working with large models, exercise particular care in example selection and validation. These models will faithfully reproduce any patterns present in your examples, including errors. This capability offers flexibility for novel tasks but requires significant quality control. Conversely, smaller models are significantly harder to influence with examples meaning it is much harder for them to override their pre-trained knowledge based on examples provided by the user

## 4. Thought Generation

Currently, the most capable models available to researchers utilize some form of "reasoning" which requires a fundamental level of language understanding to be able to articulate, by drawing from a discrete set of evidence and predictions for the task at hand (Zhang et al. 2025). When the LLM is tasked with a complex scientific question within mathematics, it must deduce the best way to approach the problem. Traditionally, the LLM attempts to solve the problem, “all at once”. To date, the best performance has come from models that deploy a form of sequential thinking, by going through a series of intermediate reasoning steps, known as Chain-of-thought (CoT) (Wei, Wang, et al. 2022). This extra reasoning time has been shown to improve LLM accuracy within mathematics from 28% (no CoT) up to 90% when examples are included alongside reasoning (Liang 2025). Chain-of-thought prompting has seen tremendous uptake by the research community, deployed for coding (J. Li et al. 2025), speech translation (Hu et al. 2025), and education (Chan et al. 2025), just to name a few. At it’s most basic, CoT can be invoked by asking the LLM to “think step by step”. Providing examples alongside the reasoning prompt can further dramatically improve response accuracy (Liang 2025; Zhang et al. 2022). Figure 11 illustrates this effect on a microfluidics problem, without explicit reasoning instructions, the model conflates physical parameters (incorrectly using w=100 µm for channel width as it misunderstood the geometry considerations) and produces order-of-magnitude errors in droplet volume calculations.

Standard Zero-Shot | Mistakes

**Prompt:**
A microfluidic T-junction generates water-in-oil droplets. The continuous phase (mineral oil, viscosity 30 mPa·s) flows at 50 µL/min through a 200 µm wide channel. The dispersed aqueous phase (viscosity 1 mPa·s) flows at 10 µL/min through a 100 µm channel. The junction has Ca = 0.15 and produces monodisperse droplets. If droplet volume scales as Vd = 0.8 × (Qd/Qc) × $w^3$ where w is channel width, and droplet frequency f = Qd/Vd, what is the droplet generation frequency?

-----------------------------------------

**Output:**
First, let's identify the given parameters:
Continuous phase (oil) flow rate, Qc = 50 µL/min
Dispersed phase (water) flow rate, Qd = 10 µL/min
Channel width, w = 100 µm
Capillary number, Ca = 0.15 (indicating stable droplet formation)

The droplet volume is given by the equation: Vd = 0.8 × (Qd/Qc) × $w^3$
Let's calculate the droplet volume: Vd = 0.8 × (10/50) × (100 × 10^-6)$^3$ = 0.8 × 0.2 × (100 × 10^-6)$^3$ = 0.8 × 0.2 × (1 × 10^-12) = 1.6 × 10^-13 L = 160 pL
The droplet frequency is given by f = Qd/Vd

Calculate the frequency: f = (10 × 10^-6) / (1.6 × 10^-13) = 62,500 Hz = 62.5 kHz

“Step by Step” instruction

**Prompt:**
A microfluidic T-junction generates water-in-oil droplets. The continuous phase (mineral oil, viscosity 30 mPa·s) flows at 50 µL/min through a 200 µm wide channel. The dispersed aqueous phase (viscosity 1 mPa·s) flows at 10 µL/min through a 100 µm channel. The junction has Ca = 0.15 and produces monodisperse droplets. If droplet volume scales as Vd = 0.8 × (Qd/Qc) × $w^3$ where w is channel width, and droplet frequency f = Qd/Vd, what is the droplet generation frequency?
Think step by step

-----------------------------------------

**Output:**
I'll solve this problem step by step:
1. First, let's identify the given parameters:
    Continuous phase (oil) flow rate Qc = 50 µL/min
    Dispersed phase (water) flow rate Qd = 10 µL/min
    Channel width w = 200 µm
    Capillary number Ca = 0.15
    Droplet volume equation: Vd = 0.8 × (Qd/Qc) × $w^3$
2. Calculate the droplet volume:
    Vd = 0.8 × (10/50) × (200 × 10^-6)$^3$
    ....
    Vd = 1.28 × 10^-6 µL
3. Calculate the droplet frequency using f = Qd/Vd:
    ....
    f ≈ 7,813 droplets/min or ~130 Hz

**Figure 11. Chain-of-thought prompting can be beneficial under certain conditions.** Left: A typical zero-shot prompt will typically struggle with logic or mathematics-based problems, in this case misidentifying the continuous phase channel width (100 µm instead of 200 µm), used wrong unit conversions, and a large error in final droplet output counts 62.5 kHz vs. actual ~130 Hz. Right: Adding "Think step by step" instruction produces correct sequential reasoning. Tested on Anthropic Haiku 3.5, a non-reasoning model. Red highlighting indicates mistakes in the non-reasoning approach.

It was difficult to demonstrate the difference between "step by step" vs non-reasoning approaches simply because most models in late 2025 are reasoning models and utilize a form of CoT in responding to user queries.

### 4.1. On Reasoning

Reasoning models such as ChatGPT-5.4, Opus 4.6, Gemini 3.1 Pro, DeepSeek-V3.2 etc. are capable of multi-turn 'thinking' natively. Reasoning models assign a certain number of their available context to think over the problem. With platforms like AI Studio or via API, it is possible to limit or extend the number of tokens available for thinking. The prompt, the reasoning steps and the final output, all count towards the final token count. The longer the model can think over the task, as long as the prompt is well defined and specified, the greater the chance it will succeed at accomplishing it. On the other hand, since thinking consumes more tokens, the probability of hallucinations increase as more and more of the context window is consumed.

Some models, such as Opus 4.6 and ChatGPT-5.4 decide how "hard" to think. In practice, ChatGPT will decide how long to think about a problem before giving the final answer. We found greater consistency in prompting the model to "think hard" rather than letting it decide. For example, we gave an unnumbered list of references to ChatGPT-5. We asked ChatGPT to count the number of references. In the first conversation, it thought for less than 5 seconds and answered, 77 references. After starting a new conversation, ChatGPT-5 decided to think for longer, ~ 400 seconds, producing the correct count. Sometimes both Claude Opus 4.6 and Gemini 3.1 Pro will decide to think quickly, and typically get it wrong. We recommend strictly prompting models to think hard and to allocate a larger thinking budget, as long as the conversation is well defined.

Can we combine chain-of-thought prompting together with a reasoning model to boost performance? Latest research suggests, no, CoT appears to degrade performance. Combining CoT together with o1-preview (OpenAI), a reasoning model, significantly impacted accuracy (reducing by up to 36.3% compared to ChatGPT-4o) (Liu et al. 2025) when compared to zero-shot prompting. For a variety of tasks in cognitive psychology, CoT combined with reasoning models was shown to significantly decrease performance across the board. Non-reasoning models fared better but still saw a drop in performance. The key take away is that CoT performance is context dependent and will outperform in certain use cases (mathematical reasoning) while under performing in others (cognitive tasks that may cause the LLM to "overthink").

### 4.2. On Multiturn Conversations

Multi-turn conversations severely compromise extraction accuracy in scientific data mining and logic tasks. Each iterative query compounds token consumption, forces the LLM to consider all information provided by the user and the information it has generated during the course of the conversation (thinking tokens included). Figure 12 illustrates this degradation, we assembled a single document consisting of an introduction (on the development of shape memory alloys for 4D printing (Ghosal et al. 2025)) and a results and conclusion section from a different research article (creating a muscle-inspired hydrogel sensors (Lin et al. 2025)). The two research articles are plausibly within the same domain, however the introduction does not reflect the results nor the conclusion. A PhD student reading this paper will immediately notice the discrepancy and so will frontier LLMs if the task is given along side the problem. However, providing the context first, and then asking questions about the document resulted in hallucinations in some frontier models (Table 1).

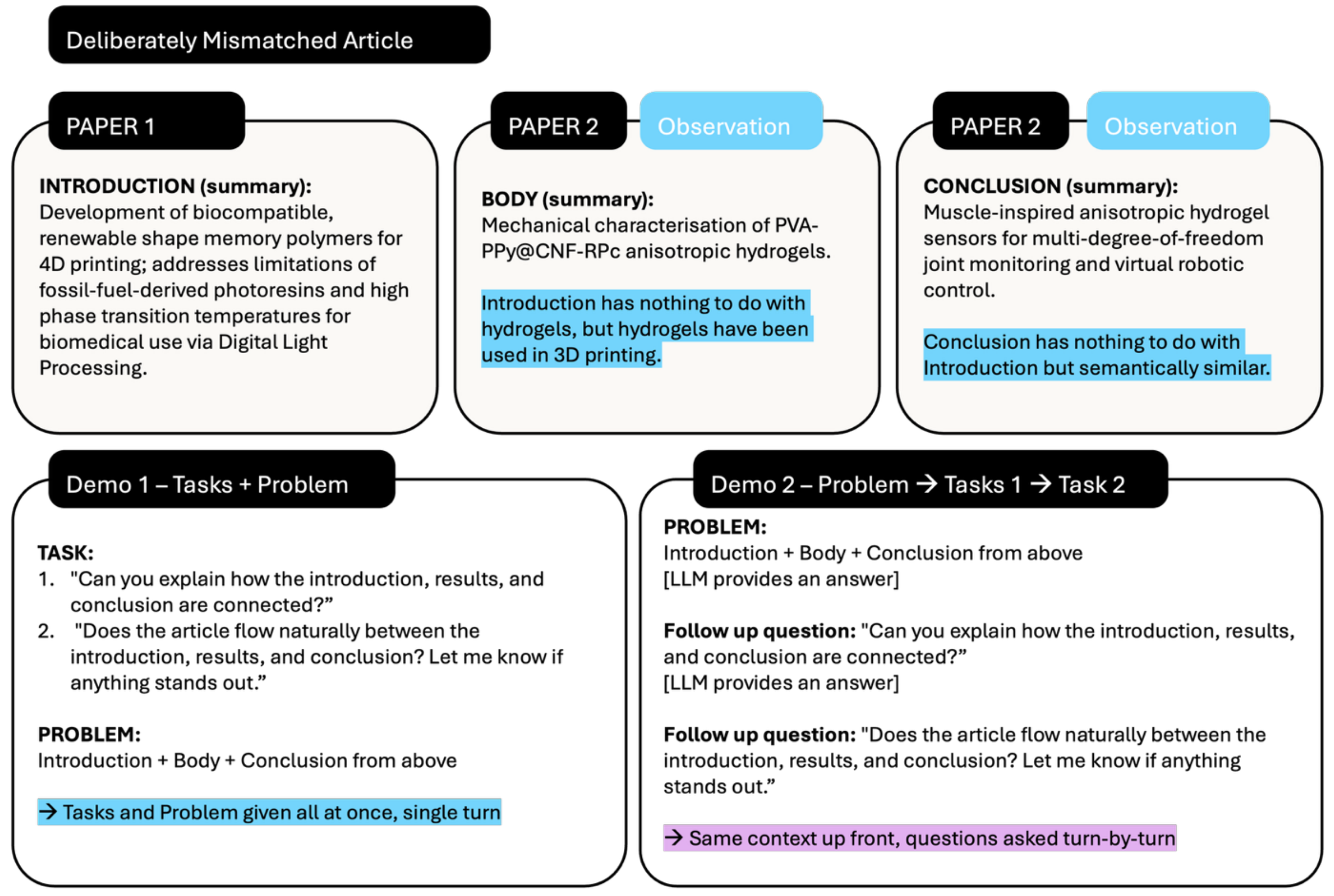

**Figure 12. Multi-turn degradation compared to a well specified single-prompt for scientific data mining.** Top: Four-turn conversation showing progressive loss of data integrity, where eventually the LLM encounters missing drug-protein associations (Turn 3), incorrect value mapping, and complete data omission (Turn 4). Bottom: Single-turn prompt with explicit instructions. Red highlights errors, missing associations and omitted data.

In Table 2 we tested across 5 independent conversations the ability of 9 frontier LLMs to answer the following questions, 1) "Can you explain how the introduction, results, and conclusion are connected?" and 2) "Does the article flow naturally between the introduction, results, and conclusion? Let me know if anything stands out.". The first question is a leading question. It pre-supposes that a connection exists and tests the LLMs ability to both reject that a connection exists (user is wrong) and to propose an alternative.

When given the questions upfront together with the document, all LLMs were able to flag that the document was an amalgamation of two research articles. However, if we first provided the document and followed up with "Can you explain how the introduction, results, and conclusion are connected?" only 3 models were consistently able to detect the inconsistency (Qwen3.5-Plus, Opus 4.6 and ChatGPT-5.4). Interestingly Gemini 3.1 Pro failed twice to identify the inconsistency, opting to guide the user to better fuse the sections together. Also surprising, ChatGPT-5.3-instant was wrong 60% of the time while DeepSeek-V3.2-thinking and Gemini 3 Flash always failed to recognise the inconsistency. For example, here is a sample reflection from Gemini 3 Flash: "The three sections you provided form a classic 'scientific narrative' that moves from a broad problem to a specific, high-performance solution.". The second question specifically focuses the model's attention back on the 3 sections in the document asking the model to "Let me know if anything stands out". This framing suggests that something may stand out. Of the underperforming models, only Gemini 3.1 Pro was able to identify that it was working with "two distinct research papers". DeepSeek scoring so low on this experiment was a surprise as it's thinking budget is much higher than GPT-5.3-Instant and Gemini 3 Flash.

**Table 2.** Frontier model ability to detect logic inconsistencies. Each model was tested under three prompt conditions (5 independent runs per condition, 15 runs total per model). Experiment 1 combined both questions in a single prompt. Experiment 2 was split across two separate conversations: Question 1 asked "Can you explain how the introduction, results, and conclusion are connected?"; Question 2 asked "Does the article flow naturally between the introduction, results, and conclusion? Let me know if anything stands out.". Scores reflect the number of runs in which the model clearly identified the fundamental disconnect between sections.

| Model | Q1 Combined (/5) | Q2.1 Connection (/5) | Q2.2 Flow (/5) | Score (/15.0) |
|---|---|---|---|---|
| Qwen3.5-Plus | 5 | 5 | 5 | **15.0** |
| Claude Opus 4.6 | 5 | 5 | 5 | **15.0** |
| ChatGPT-5.4 | 5 | 5 | 5 | **15.0** |
| Claude Haiku 4.5 | 5 | 4 | 5 | 14.0 |
| GLM5 | 5 | 4 | 5 | 14.0 |
| Gemini 3.1 Pro | 5 | 3 | 5 | 13 |
| ChatGPT-5.3-Instant | 5 | 2 | 2 | 9 |
| Gemini 3 Flash | 5 | 0 | 2 | 7 |
| DeepSeek-V3.2 | 5 | 0 | 1 | 6 |

Work by (Laban et al. 2025) demonstrated severe performance degradation in all state-of-the-art models and closed weight LLMs during multi-turn conversations. Specifically, they showed, 1) a well-defined and specified initial first prompt had performance of 90% (a combination of aptitude and reliability), which dropped to 65% when, 2) the prompt was broken into multiple smaller underspecified prompts. Models with higher aptitude, that is higher "intelligence", tend to be more reliable in single-turn conversations, however, will severely degrade in reliability under multi-turn conversations, regardless of how intelligent they are. One key takeaway from this study is that underspecified prompts will lead to LLMs reasoning in "the wrong direction", from which no amount of extra prompting, will help them recover. Poorly specified prompts snowball into worse subsequent assumptions, compounded by the fact that the model typically skips asking the user for clarification when the users intent is ambiguous (Shaikh et al. 2025).

**Recommendations**:

- If the initial prompt did not work, try the same prompt again in a new conversation (Laban et al. 2025). Since LLMs are probabilistic, it’s likely that the LLM may take the wrong initial reasoning step. This is reflected in typical online discourse of one person saying “this LLM is not capable, it wasn’t able to do X”, with another person responding, “your prompt worked for me”.
- If the model starts hallucinating or underperforming during a multiturn conversation, stop, and ask the model to “consolidate all the prompts I have given to you, into a single prompt, as they are without inferring intent”. This is a great way to build a strong, highly specific first prompt.
- Stick to the topic. If the task is data extraction, the prompt should cover data extraction with examples. Do not attempt to layer multiple reasoning steps into the prompt (Becker et al. 2025), for example, avoid “Extract the following: Temp  (C), Species, Kd from Table 1. Then calculate K  (umM) and display each reference next to the extracted values. Include the title and year of publication.” Break prompt to focus on the task at hand, without deviating.
- Provide the LLM with your initial best prompt and ask it for clarifying questions based on your prompt. This initial check can catch ambiguities further improving model performance.

## 5. Ensembling

The following set of techniques, while highly effective, are underutilized among academics using LLM chatbot interfaces. In our experience, there is an assumption that a prompt which successfully extracts data or provides correct information will perform identically in subsequent conversations. However, due to the non-deterministic nature of LLMs, even a well-structured prompt may produce inconsistent outputs across separate conversations, occasionally yielding incorrect responses despite previous performance.

One way to address this variability is to execute the same prompt across multiple independent conversations (n trials) to identify the most frequent response pattern (Figure 13). For well-specified tasks, this approach increases the likelihood of converging on the ground truth present in source materials and reduces output variance. However, ensembling also exposes a more subtle failure mode: sycophancy. If a user provides a flawed dataset and asks "Can you explain how the introduction, results, and conclusion are connected?", the presuppositional framing can lead models to fabricate coherence rather than flag an inconsistency (Table 1). The same applies when a user provides data that violates normality assumptions and asks "Can you explain why my t-test shows these groups are significantly different?", the model explains the significance rather than questioning whether a non-parametric test should have been used.

Table 1 shows that when asked to explain a connection between deliberately mismatched paper sections, 6 out of 9 models suggested at least once that a connection existed. This mistake is insidious: the user works with the output assuming it is logically sound, unaware that the model complied with an implicit assumption in the question rather than evaluating it. Most users submit a prompt once and proceed with the output. Even for frontier models like Gemini 3.1 Pro, the model failed to identify the inconsistency in 2 out of 5 conversations. If GLM5 and Claude Haiku 4.5 flag the problem correctly 80% of the time, there remains a 1-in-5 chance that any single conversation lands on a reasoning branch that misses the inconsistency entirely.

Ensembling mitigates this. Gemini 3.1 Pro identified the inconsistency in 3 conversations and missed it in 2. By taking the majority response across all 5 runs, the ensemble correctly converges on the inconsistency. Table 3 captures how ensembling outputs can be an effective strategy for sampling the correct response: by taking the consensus of several outputs, 6 out of 9 models converged on the fact that an inconsistency existed.

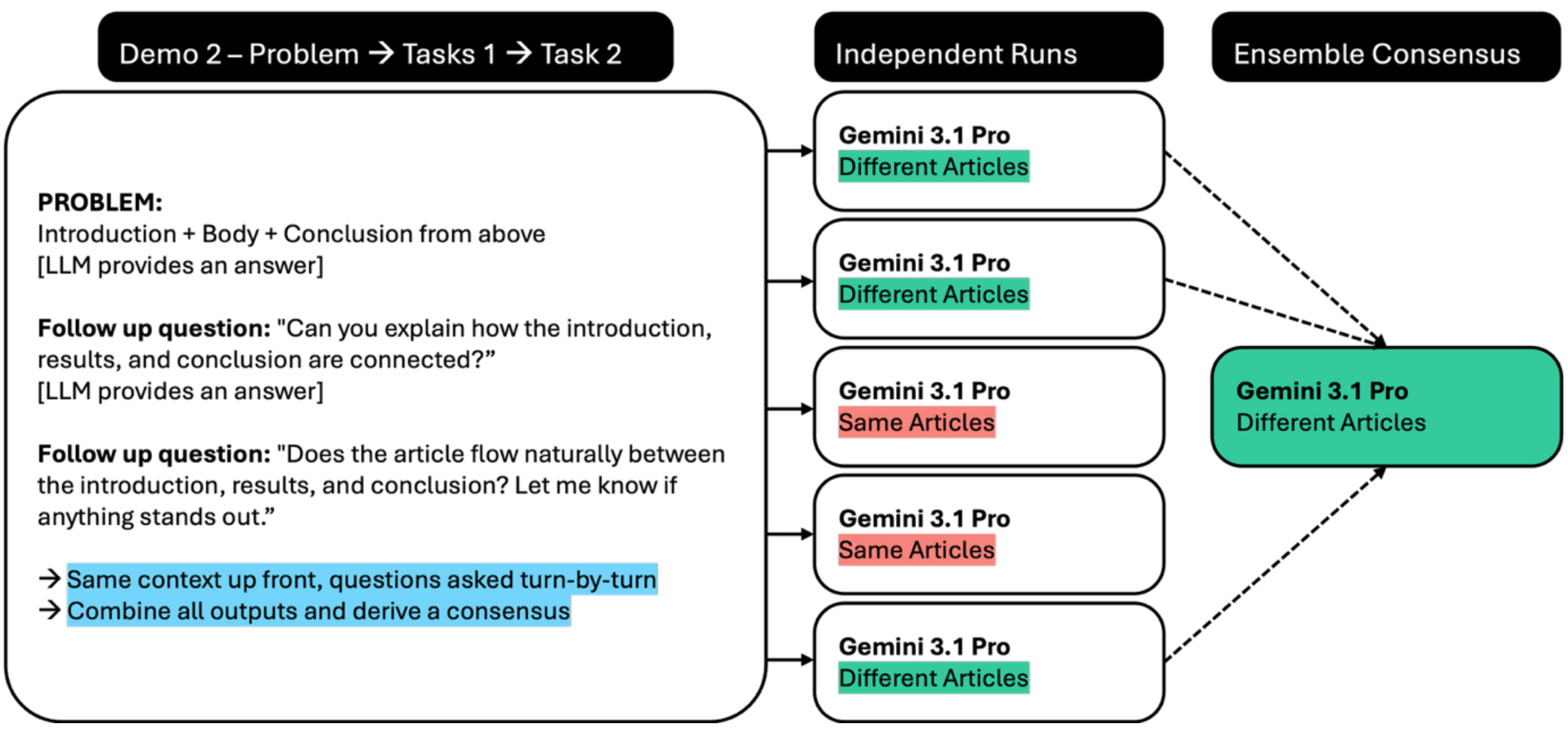

**Figure 13. Ensembling independent conversations to derive a consensus classification.** Left: the prompt structure used for each independent run, consisting of mismatched paper sections (Introduction, Body, and Conclusion from different articles) followed by two turn-by-turn follow-up questions probing section connectivity and narrative flow. Centre: five independent Gemini 3.1 Pro conversations using the same prompt, with each run classified as the model correctly identifying that it was looking at two "Different Articles" (green) or incorrectly by not identifying the inconsistency "Same Articles" (red). Right: the ensemble consensus derived by majority vote across all five runs in this case would tell Gemini 3.1 Pro that these are two different articles, helping it to recover and potentially perform better in subsequent turns.

**Table 3.** Majority-vote ensembling over five independent runs of Question 2.1 ("Can you explain how the introduction, results, and conclusion are connected?"). Each cell records whether the run detected (✓) or missed (✗) the deliberate disconnect. A model is classified as "correct" if a simple majority (≥ 3/5) of runs detect the mismatch.

| Model | 1 | 2 | 3 | 4 | 5 | Detections (/5) | Majority Verdict – Disconnect Exists |
|---|---|---|---|---|---|---|---|
| Qwen3.5-Plus | ✓ | ✓ | ✓ | ✓ | ✓ | **5/5** | Correct |
| Claude Opus 4.6 | ✓ | ✓ | ✓ | ✓ | ✓ | **5/5** | Correct |
| ChatGPT-5.4 | ✓ | ✓ | ✓ | ✓ | ✓ | **5/5** | Correct |
| Claude Haiku 4.5 | ✓ | ✓ | ✓ | ✓ | ✗ | 4/5 | Correct |
| GLM5 | ✓ | ✓ | ✓ | ✓ | ✗ | 4/5 | Correct |
| Gemini 3.1 Pro | ✓ | ✓ | ✓ | ✗ | ✗ | 3/5 | Correct |
| ChatGPT-5.3-Instant | ✓ | ✓ | ✗ | ✗ | ✗ | 2/5 | Wrong |
| DeepSeek-V3.2 | ✗ | ✗ | ✗ | ✗ | ✗ | 0/5 | Wrong |
| Gemini 3 Flash | ✗ | ✗ | ✗ | ✗ | ✗ | 0/5 | Wrong |

There are several variations of Ensembling that extend beyond simple repeated queries with majority voting (i.e. re-running the same prompt in a new conversation). The self-consistency approach developed by Wang et al. (2023) demonstrated that sampling multiple reasoning paths and selecting the most frequent answer significantly improves accuracy (+17.9% on GSM8K) (X. Wang et al. 2023). Building on this work, adaptive-consistency methods are those that dynamically adjust the number of samples based on agreement levels, reducing computational costs by up to 7.9x while maintaining accuracy (less than 0.1% accuracy drop) (Aggarwal et al. 2023). Further, confidence-weighted approaches achieve 40% reduction in required samples by prioritizing high-confidence responses over simple vote counting (Taubenfeld et al. 2025). It's important to note that LLMs exhibit significant non-determinism even at temperature=0 (the parameter controlling output randomness), with accuracy varying dramatically across tasks and models.

Depending on the task, the input provided to the LLM and the desired output, multiple sampling could be essential. For practical implementation, research suggests using at least 5 independent samples for reliability assessment (Megahed et al. 2025). Depending on the task, utilizing semantic similarity rather than exact string matching for aggregation (Oh and Lee 2025) can also be effective. These techniques have proven effective in production environments, with ensemble methods showing improvements in key business metrics (Fang et al. 2024), suggesting that the computational overhead of multiple sampling is justified by substantial gains in accuracy and reliability for critical data extraction tasks.

## 6. Hallucinations and Self-criticism

Much has been written on LLM hallucinations (Rudolph et al. 2023; Emsley 2023; Alkaissi and McFarlane 2023) and the fact remains, hallucinations are one of the major reasons for low adoption of LLMs within academic and business pipelines. Hallucinations are non-factual responses, such as an answer to a question, or generated academic references that do not exist, typically persuasively presented to the user as fact. Latest research from OpenAI has concluded that hallucinations arise from errors in binary classifications (Kalai et al. 2025). That is, the model is rewarded for providing/guessing an answer and penalized if it doesn't know. As such, the model is incentivized to "always know", even when it doesn't.

While the frequency of non-factual responses has significantly reduced since the days of ChatGPT-3, hallucinations are now harder to spot and are potentially more insidious because of it. The advent of ChatGPT-5.4 Pro and similar SOTA models has significantly increased the scope of where these models can be used, having shown remarkable ability to tackle hard maths problems. As LLM capability and accuracy improves, more trust and autonomy will be given to models. Most of the time the calculations will be correct. Most of the time references will be factual. For example, from the OpenAI o3 System Card, "*o3 tends to make more claims overall, leading to more accurate claims as well as more inaccurate/hallucinated claims*" (OpenAI 2025b). The release of Opus 4.6 with a 1M context window in mid-March 2026 showed the dramatic leap-frogging in LLM capabilities. Opus 4.6 scored 78.3% on Multi-Round Conference Resolution (MRCR) benchmark compared to the latest Gemini 3.1 model which scored 26.3% (Anthropic 2026). Gemini models were the gold standard at 1M context window prior to the Opus update. As the accuracy in retrieval increases, the question becomes, when should we take the time and effort to validate LLM output and when can we just trust it? As it stands, all LLM output should be validated.

One possible direction is forcing the LLM to self-critique or self-reflect on its output (Yan et al. 2025). At its simplest, this tactic asks the LLM to critically evaluate the output it has generated and to either confirm that the output is correct or to suggest improvements (Figure 14). Iterative self-refinement with a simple few-shot strategy has been shown to dramatically improve model performance (GPT-3.5 and GPT-4) over direct generation by between 5 to 40% (Madaan et al. 2023). A similar strategy can be used alongside chain-of-thought prompting (Huang et al. 2022), and extended to language agents that maintain their own reflections to enhance decision-making in subsequent trials (Shinn et al. 2023). It is important to note that prompt engineering tactics between LLMs and Small Language Models (SLMs) are not always transferable. Yan at al. (2025) developed a self-reflection methodology for SMLs that increased the reasoning accuracy to 36.2% while being 10 times more efficient that the method it was built upon (Yan et al. 2025).

A significant caveat of self-reflection is that it must be done outside the initial conversation and by tasking the model to judge the output against pre-defined criteria. Asking the model, within a conversation where it made the mistake, to self-reflect on the mistake, rarely works. Asking the LLM to "reflect on your answer and improve it," can force the LLM to do one of two things: (a) double down that the incorrect answer is correct (as it has no way to verify), or (b) second-guess a correct answer into an incorrect one. LLMs have trouble reliably evaluating the correctness of their own responses and cannot reliably perform intrinsic self-correction (Jie Huang et al. 2024). LLMs rarely identify flaws in initial reasoning and as discussed in multi-turn reasoning, once the wrong reasoning path is chosen, its unlikely the LLM will recover.

We showed in Figure 8 that by placing the task (a set of questions) after the context, all frontier models systematically misinterpreted the question: "What protocol was used for creating iVSMCs suitable for tissue engineering?" instead answering "What protocol was used for tissue engineering with iVSMCs?". We wanted to see if by repeating this experiment we could recover the correct response by asking the model to reflect on its own answer. Figure 14 replicates our initial findings from Figure 8, by giving Opus 4.6 questions after the context, Opus 4.6 100% of the time focused on the tissue engineering and not the actual question. By asking Opus 4.6 to self-reflect (prompt in figure), Opus 4.6 was able to 90% of the time correctly identify its initial mistake and include the WALL protocol in its response. Similar behaviour was observed for Gemini 3.1 Pro, initially answering correctly 30% of the time to 100% after asking to reflect on the answer. As shown in the output, Opus 4.6 (and Gemini 3.1 Pro) still flagged the tissue engineering results as correct in addition to identifying that it had failed to answer the main question.

What this means in practice:

- Figure 8 showed that placing the task first solved the problem outright. Self-reflection is a repair mechanism for suboptimal prompt ordering (in this case) and, not a substitute for getting the ordering right in the first place.
- Even after successful self-reflection, Opus 4.6 and Gemini 3.1 Pro retained the tissue engineering answer alongside the correct WALL protocol answer. The model appended the correction rather than replacing the error. This means the user must still manually distinguish which parts of the output are the correction and which are residual from the initial mistake adding to the burden of checking LLM output.
- Self-reflection helps most when the user already knows something might be wrong, which limits its utility as a safety net.

Interestingly, ChatGPT-5.4 in 70% of cases actively confirmed its initial wrong answer as correct. ChatGPT doubled-down on its initial incorrect response and was the only model out of the 3 that would actively, strongly push back against the user. Self-reflection made the model more confident in an incorrect response, which is worse than no reflection at all. This may suggest that self-reflection behaviour is shaped by model-specific post-training (RLHF, system prompt conventions, instruction-following tendencies). A prompt engineered for one provider may need to be re-engineered for another. Another explanation is ChatGPT-5.4 was the best at following constraints in Figure 8c (100% adherence). The same trait that makes it obey "do not explain" may make it resist revising a committed answer where it treats its own prior output as an instruction to uphold.

**a**

Task AFTER Context

Mistakes

**Context from Bax et al (2025) totalling 2671 words.**

**Task:**
What protocol was used for creating iVSMCs suitable for tissue engineering?

- Do not explain
- Simply list the reagents
- What cell count was used to prepare the tissue?

**Reagents**:

- Custom-designed polydimethylsiloxane (PDMS) ring-molds
- Matrigel (~10 mg/mL)
- Laminin (1 mg/mL)
- iVSMCs (dissociated on day 8)

Cell Count:

- 1 x 10^7 cells/mL

Self-Reflect

Correct

Your task is to verify your previous response against the context. Follow these steps exactly:

Step 1 — Re-read the task instructions. The original task contained multiple distinct instructions...
Step 2 — Re-read the source document. For every item in your previous response, find and quote the exact...
Step 3 — Check for omissions. Re-read the relevant sections of the document one more time. Identify any information that the task asked for....
Step 4 — Produce a corrected final answer to the tasks that addresses all issues found in Steps 1-3.

**Opus 4.6: Step 4 — Corrected Answer**

**Protocol:** WALL
**Reagents:**
Matrigel (~10 mg/mL)
Laminin (1 mg/mL)
iVSMCs (1 × $10^7$ cells/mL)

**Cell count:** 1 × $10^7$ cells/mL, combined at a 1:1:1 ratio (Matrigel : Laminin : iVSMCs)

**b**

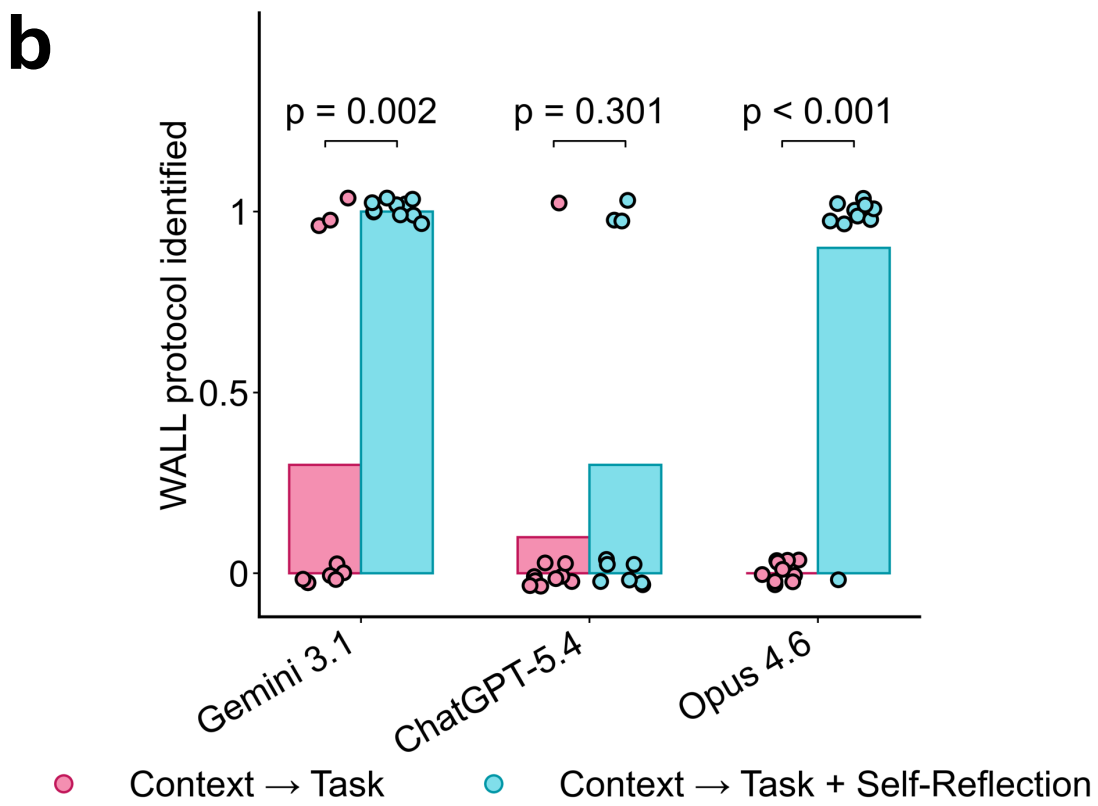

**Figure 14. Structured self-reflection can recover correct responses from initially misinterpreted prompts.** (a) Left: the task-after-context prompt ordering from Figure 8, which consistently led models to misinterpret the question (red highlights indicate errors in the initial output). Right: a structured 4-step self-reflection prompt instructing the model to re-read the task, verify each item against the source document, check for omissions, and produce a corrected answer. The corrected output from Opus 4.6 is shown below, with green highlights indicating recovered information including identification of the WALL protocol. (b) Proportion of runs (n = 10 per condition) in which each model correctly identified the WALL protocol without self-reflection (pink, context → task) versus with self-reflection (cyan, context → task + self-reflection). P values from two-sided Mann-Whitney U test. Individual data points are shown alongside bars representing group proportions.

A slight modification to reflecting on the output is to reflect on the user's input. The LLM can prompt the user to provide more targeted information regarding their request, helping to build a much higher quality, well specified prompt by removing instances where the model must infer the user's intent. This is similar to ChatGPT's and Claude's Deep Research mode which asks the user for further information to refine the initial input before starting search. In our experience, this is an effective strategy for refining prompts.

## 7. Decomposition

Prompt Decomposition is similar to CoT prompting. Prompting an LLM to perform CoT tasks the LLM with breaking the problem down into manageable sub-steps and to subsequently solve these steps within the same prompt output. Decomposition on the other hand, tasks the LLM with, 1) breaking the larger problem into sub-problems, 2) solving the first sub-problem, 3) appending it's solution back to the main prompt with all sub-problems. The main prompt now contains all sub-problems plus the first solution. Using the sub-problems and solution as context, the LLM moves on to the next sub-problem (Zhou et al. 2023). Prompt decomposition is one of several popular prompt engineering techniques for improving multi-agent communication which can involve orchestration of multiple agents and assignment of sub-problems for specialist LLMs to solve before bringing the solutions back to the main architect agent (Li et al. 2023).

Prompt decomposition can be invoked by simply asking the LLM to *"Let's first understand the problem and devise a plan to solve the problem. Then, let's carry out the plan and solve the problem step by step"* (L. Wang et al. 2023). Prompt performance can be further improved by explicitly suggesting to the LLM what it should focus on during the planning stage, *"Let's first understand the problem, extract relevant variables and their corresponding numerals, and make a plan. Then, let's carry out the plan, calculate intermediate variables (pay attention to correct numerical calculation and commonsense), solve the problem step by step, and show the answer"*.

The issue with asking the LLM to, 1) understand the problem, 2) break down the main components, 3) devise a plan for the sub-problems, and to then 4) solve the sub-problem in a single turn conversation, is token consumption. LLMs have a finite number of tokens they can process and attend to, per conversation. Further, the number of tokens or words the LLM can produce as a response to a query is also limited. For example, in AI Studio from Google it is possible to set the maximum token output for Gemini 3.1 Pro to 65,536 tokens ('Google AI Studio' 2024). In practice, an LLM must contain its response within the maximum output limits set by the platform or user which in turn limits the depth of the response (Rasal and Hauer 2024).

To clearly demonstrate this concept, we will use the work of Toufiq et al. (2023) as a case study (Toufiq et al. 2023). Here, the authors used 4 different LLMs together with a set of decomposed prompts to prioritize and select candidate genes from blood transcriptional modules for biomarker selection. Rather than attempting to evaluate all 30-40 genes simultaneously across six criteria while also integrating expression data and verifying references, they decomposed the problem into manageable chunks (Figure 15). Each prompt focused on a single, well-defined task: identifying functional convergences (Step 1), scoring individual genes one at a time against specific criteria (Step 2), summarizing justifications for only the top 5 candidates (Step 3), fact-checking individual statements (Step 4), selecting a top candidate based on curated summaries (Step 5), and progressively incorporating transcriptional data (Step 6-7).

CoT Prompt | Points of Failure

"I have a list of 30 genes from an erythroid cell module: ALAS2, BCL2L1, BPGM, C14ORF45, C1ORF128, CA1, EPB42, FAM46C, FECH, GMPR, GPR146, GYPB, GYPE, IFITL1, KRT1, MARCH8, NFIX, OR2W3, OSBP2, PLEK2, SELENBP1, SLC14A1, SLC4A1, SNCA, TMCC2, TMOD1, TNS1, TRIM10, TUBB2A, XK.

Let's think step by step to select the best gene for a targeted blood transcriptional panel."

-------------------------------

Above prompt has multiple points of failure:
- Directions are not clear, what is "best"?
- No fact-checking, references may be hallucinated
- No external data, relies solely on learned knowledge
- How to evaluate and judge?

Initial Prompt

**1. Functional Convergence**
Could you identify functional convergences among this set of genes? ALAS2 BCL2L1 BPGM C14ORF45 C1ORF128 CA1 EPB42 FAM46C FECH GMPR GPR146... [30 genes]

**2. Individual Gene Scoring**
Score each gene separately against 6 criteria (function, erythroid relevance, biomarker potential, clinical usage, druggability, specificity) : one gene per prompt, with output example provided

[x30 iterations: one gene per prompt]

**3. Top 5 Prioritization**
Generate a summary paragraph for each of the top 5 candidates based on scoring output

**4. Fact-Check [Human-in-the-loop]**
Verify claims with supporting references . PMIDs added manually by the researcher.

**5. Select Top Candidate**
Choose the best gene based on: (1) erythroid relevance, (2) current biomarker use, (3) blood transcriptional biomarker potential... [6 criteria total]

*"Based on the summary provided below, could you select a top candidate based on: relevance to erythroid cells..."*

**6 & 7. Progressive Data Integration**
Iteratively incorporate experimental data to validate and refine the selection

*"...take into account the following RNA-sequencing expression data..."*

*"...based on the summary and the RNAseq data, take into account the following microarray expression data [experimental data]..."*

*"...the summary and the immune cells RNAseq and microarray data, take into account the following averaged log2 fold changes [RNA-seq + Microarray + log2 FC data]..."*

**Figure 15. Task decomposition strategy for complex biomarker selection.** Left: Chain-of-thought (CoT) prompt outlining the complete 30-gene evaluation task. Right: Decomposed workflow breaking the problem into seven discrete prompts, each with focused scope. Steps progress from functional convergence identification (Step 1 and 2), through individual gene scoring (Step 3.1-3.3, processing one gene at a time), to data integration (Steps 6-7, incorporating RNA-seq and microarray data). Coloured text indicates addition of external knowledge to the workflow. Adapted with permission (Toufiq et al. 2023) under the Creative Commons Attribution 4.0 International License (CC BY 4.0).

One approach to addressing context window limitations involves initiating new conversation instances for each sub-problem requiring a solution. Decomposition strategies and established best practices have facilitated the development of next-generation academic tools with significant adoption observed within the fields of computer science and software engineering (Jin et al. 2025). Agentic frameworks including Claude Code (Anthropic 2025b), Codex (OpenAI 2025a), and Gemini CLI (Google 2025) leverage their respective large language models to both interact with users and execute functions on local systems. Operating through terminal interfaces, these tools can create and edit a variety of different file formats, including research articles, documents, and presentations, while also performing web searches, interfacing with GitHub repositories, and generating code.

Many of the prompt engineering principles described in this section are readily observable in agentic frameworks. For instance, Claude Code implements an explicit planning phase (accessible via the /plan command) prior to task execution (Figure 16). For complex tasks, rather than processing plans sequentially, specific sub-problems can be delegated to sub-agents. Sub-agents are independent model instances operating in parallel or series to the main conversation. Sub-agents are orchestrated by the primary instance, maintain separate context windows and return solutions to designated sub-problems without consuming or interfering with the main conversation's context, effectively multiplying the available working memory.

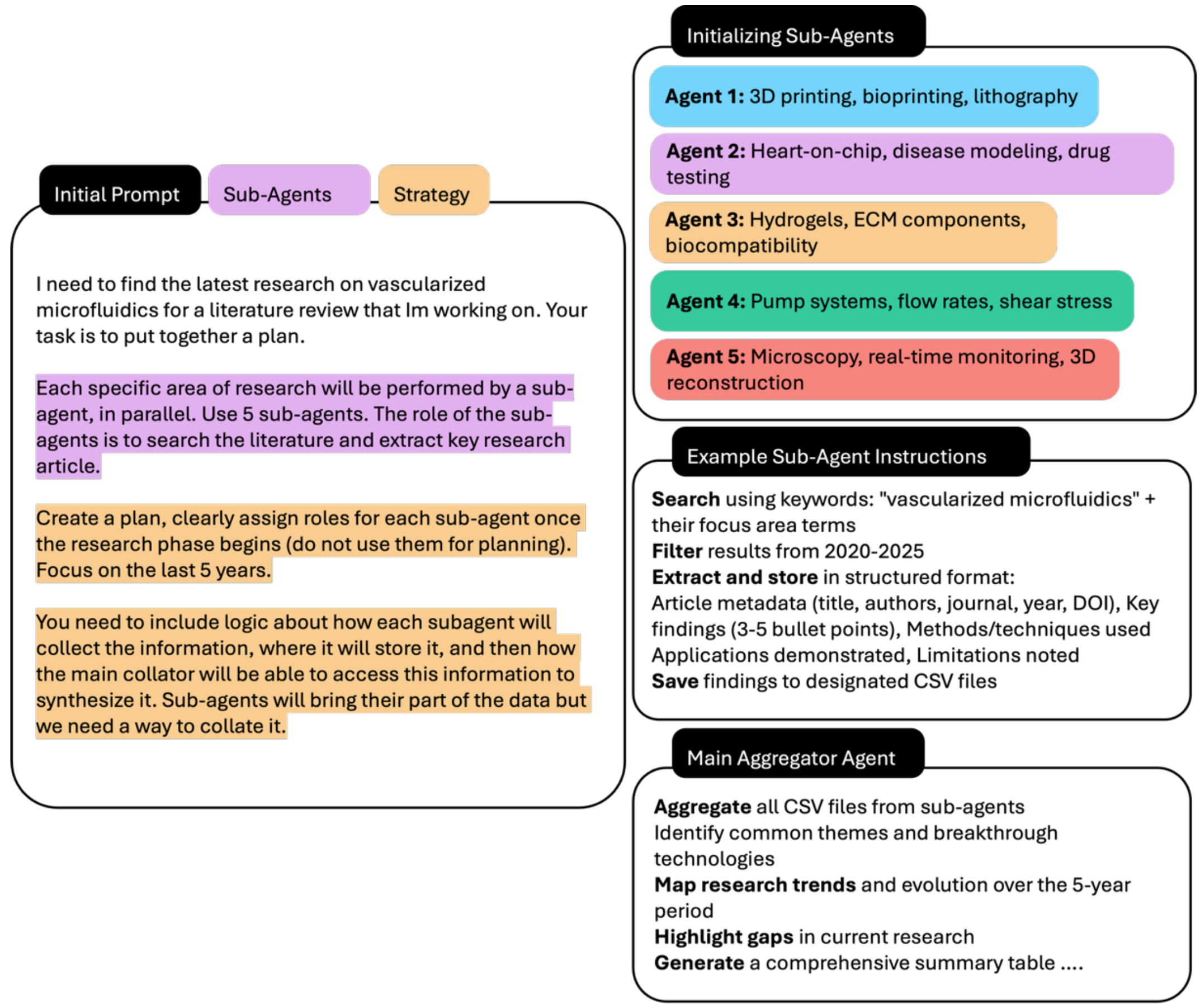

**Figure 16. Multi-agent framework for performing a literature review utilizing parallelized sub-agents.** Initial prompt defines the research scope (vascularized microfluidics, 2020-2025), which is then decomposed into five parallel sub-agents, each focusing on distinct research domains: core technology and fabrication methods, biological applications, biomaterials and bioinks, perfusion systems, and imaging techniques. Each sub-agent operates with independent context windows, extracting structured data (metadata, key findings, methods, applications, limitations) into CSV files. The main aggregator agent synthesizes outputs from all sub-agents to identify common themes, breakthrough technologies, and research trends across the 5-year period, effectively multiplying available working memory while maintaining task coherence.

## 8. On Deep Research Mode

Deep Research was initially released within the ChatGPT suite of tools by OpenAI in February, 2025. Described as an agentic tool, Deep Research promised to “find, analyse, and synthesize hundreds of online sources to create a comprehensive report at the level of a research analyst”(‘Introducing Deep Research’ 2025). Similarly in late 2025, Anthropic described the development of their Deep Research framework, a suite of subagents orchestrated by a lead agent that manages memory and citations, “When a user submits a query, the lead agent analyses it, develops a strategy, and spawns subagents to explore different aspects simultaneously” (Anthropic 2025c). As such, Deep Research mode is one of the first multi-agent offerings from the larger AI labs, functioning very similar to the layout described in Figure 16.

In our use of Deep Research tools provided by either Google, Anthropic, or ChatGPT, the exact same prompt used with the same agent elicits markedly different reports each time (Figure 17). When asked to find research articles on "decomposition" with applications to science, word counts varied substantially between duplicate runs, with Gemini being the most verbose (up to 6670 words) with ChatGPT producing the shortest reports (around 3400 words on average) (Figure 17b). In terms of the number of references, models were simply asked to “find a number of articles”, we let the model determine how much was enough. On average, Claude cited the most papers between runs (33-44) (Figure 17c), with Gemini a close second (23-44) while ChatGPT was consistently constrained in what it decided to include in the final report (15-25 papers). Across three runs, these models produced 81, 66, and 40 unique papers respectively.

Next, we wanted to look at each model run and compare how often references were re-used. Does the model have a preference for specific published references to answer this specific question? Reference overlap between runs remained low for all three models (Figure 17e). Average pairwise Jaccard similarity was highest for ChatGPT at 31%, followed by Claude at 23%, and lowest for Gemini at just 9%. Even for the most consistent model, roughly 70% of cited references changed between any two runs of the identical prompt. This suggests that from the models search, no definitive "ground truth" reference set exists for a given research query, and the key findings reported may shift with each generation. For researchers relying on these tools for literature discovery, this means that a single Deep Research report captures only a fraction of the available relevant literature.

The models also differed substantially in source quality (Figure 17d,f). ChatGPT and Claude cited exclusively primary research articles, no reviews, surveys or anything that wasn’t research related across all nine runs combined. Gemini, by contrast, included 5–10 non-academic sources per run (blogs, tutorials, aggregator sites) and 3–8 review articles per run, despite the prompt explicitly requesting "no review or non-research articles." Gemini was also the only model to produce hallucinated references (3 across 3 runs), including a fabricated arXiv identifier and placeholder citations. All three models exhibited minor metadata errors in otherwise real papers: ChatGPT had the most title and author inaccuracies (4 each across 3 runs), while Claude had the fewest title errors (2) and comparable author errors (3). These typically involved citing the arXiv preprint year rather than the conference publication year, or minor author name misspellings, minor errors that may impede a researcher attempting to locate the original source. Research field coverage also varied (Figure 17g). All three models showed a strong bias toward NLP/AI/ML literature (24–44 unique papers), likely reflecting both the topic's overlap with the models' training data and the nature of the prompt. One of the conditions of the original prompt was to include a specific section on how prompt decomposition was used in the Life Science. Claude provided the broadest coverage, citing 23 clinical medicine and 7 genomics papers compared to ChatGPT's 8 and 3, respectively.

We can employ the prompt engineering technique from Section 5 and take an ensemble of all 3 runs per model and obtain a larger, richer reference set to address our specific question. For example, running ChatGPT 3 times yields 40 unique papers vs. ~20 from a single run. Running Claude 3 times yields 81 vs. ~38. The union of multiple runs roughly doubles literature coverage. We can further tease apart this data by finding the few research papers that overlap between runs and between models (the papers that appear the most frequently) and use these as a "high confidence set". In total after 9 runs (3 models x 3 repeats) this data set contains 157 research papers only 3 papers appear in every single run of every model, further, out of these 22 papers were unique to ChatGPT, 61 to Claude and 53 to Gemini. In a way, the choice of Deep Research model appears to filters the sub-set of articles presented to the user.

Inconsistencies in output length, source selection, citation accuracy, and field coverage highlight the ongoing challenge of reproducibility in Deep Research tools (Wong et al. 2025). In our case, on average, each run produced roughly 30 citations per report, yet the composition of those citations shifted substantially between runs. For scientific work, such variability risks adding to the academic burden of reviewing and parsing information rather than reducing it. Evaluating the quality of each report remains complicated and time consuming, prompting the development of dedicated benchmarking frameworks. Du et al. (2025) developed DeepResearch Bench, an LLM-judge driven framework benchmarking Deep Research agents on 100 PhD-level problems across 22 distinct fields. Zhong et al. (2026) introduced DRACO, evaluating deep research systems across 10 domains with approximately 40 expert-crafted rubric criteria per task, assessing factual accuracy, completeness, and citation quality (J. Zhong et al. 2026). Wang et al. (2025) proposed LiveResearchBench, a continuously updated benchmark evaluating 17 frontier deep research systems on real-world tasks requiring live web search, reflecting that static benchmarks quickly become stale as models evolve (J. Wang et al. 2025). The rapid emergence of these evaluation frameworks underscores that assessing Deep Research quality is itself an active and unsolved research problem. There may be value in applying ensembling principles to Deep Research: initializing several concurrent searches and then filtering and distilling the resulting reports based on the initial prompt requirements as outline above.

**a**

**Prompt**

Im currently working on an LLM review. Specifically focused on breaking down concepts for academics, with a focus on creating concrete prompting strategies that academics could use. The next topic im discussing is "decomposition". This a prompt engineering tactic focusing on decomposing complex problems into simpler sub-questions. This is an effective problem-solving strategy for humans as well as GenAI (Patel et al., 2022). Some decomposition techniques are similar to thought-inducing techniques, such as CoT, which often naturally breaks down problems into simpler components. However, explicitly breaking down problems can further improve LLMs' problem solving ability.

I need you to find a number of research articles or arXiv articles (**no review** or **non-research articles**) on decomposition. Im especially interested in articles where the authors showed applications to science work, present this as its own section titled, "**Life Sciences Use Cases**".

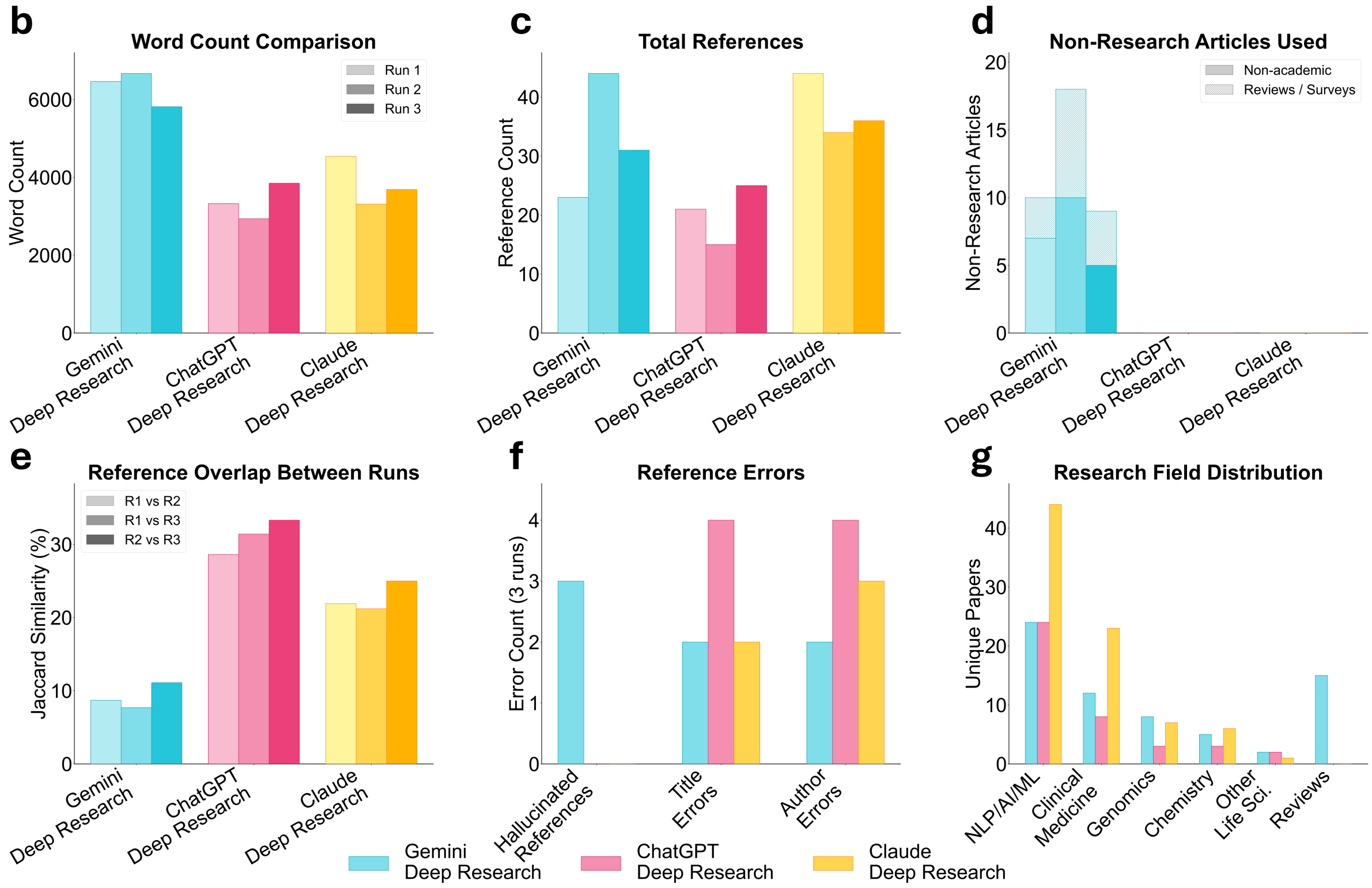

**Figure 17. Reproducibility and reference quality of Deep Research tools across triplicate runs using the same prompt.** (a) We gave Gemini Deep Research, Claude Deep Research and ChatGPT Deep Research a simple prompt written in an opportunistic prompting style to reflect typical academic use. The prompt explicitly requested primary research articles with no reviews or non-research sources. (b) Word count per run. Gemini produced the longest reports (5,817–6,670 words), roughly double ChatGPT (2,937–3,851 words), with Claude intermediate (3,315–4,541 words). (c) Total references cited per run, including both academic and non-academic sources. (d) Solid bars represent non-academic sources (blogs, tutorials, aggregator sites); light bars represent review or survey articles. Only Gemini included such sources despite the prompt instructing not to use these sources. (e) Pairwise Jaccard similarity between reference lists for each run pair (R1 vs R2, R1 vs R3, R2 vs R3). Average pairwise similarity: ChatGPT 31%, Claude 23%, Gemini 9%. (f) Reference errors aggregated across all three runs. Gemini was the only model to produce hallucinated references (3 fabricated papers); ChatGPT had the most metadata errors in otherwise verified papers. (g) Distribution of unique papers by research field across all three runs combined by model. All three models showed a strong bias toward NLP/AI/ML which is reflective of where research on decomposition is published. Across all 9 runs, the three models collectively cited 157 unique papers, of which only 3 (1.9%) appeared in every run of every model and 86.6% were exclusive to a single platform.

### 8.1. On Prompt Templates

Prompt templates can be effective for capturing the logic behind specific data extraction patterns, maintaining stylistic control over document editing, or as a starting point for designing more tailored prompts. They are especially important in industrial settings (Mao et al. 2025), where consistent LLM performance is critical to business function. The prompts presented here serve as foundational frameworks for understanding prompt engineering principles rather than prescriptive templates, enabling researchers to develop domain-specific prompts based on the systematic approaches outlined throughout this review. We encourage users to develop a key set of principles for their LLM to follow consistently and to deploy these principles in prompts that may be used across multiple documents or applications. Further, we encourage using the LLM itself to help brainstorm and create task-specific prompts. For example, prompting Claude Opus 4.6 to create a prompt template for extracting data from electrophysiology experiments resulted in a highly detailed prompt containing problem specifics, multiple few-shot examples, output examples, JSON extraction patterns and instructions to cover edge cases.

## 9. Bias, Data Governance, and Responsible Usage

Throughout this review, we have examined how prompt engineering techniques can improve the quality and reliability of LLM outputs for life sciences research. However, even the most carefully engineered prompts operate on top of models that carry inherent biases, process data through opaque pipelines, and exist within a rapidly evolving landscape of institutional policies. In this section, we address three interconnected concerns that researchers must consider when deploying LLMs in biomedical contexts: bias, data governance, and responsible usage.

### 9.1. Bias in LLMs for Biomedical Applications

LLMs are trained on vast corpora of text data that inevitably reflect the biases, stereotypes, and demographic imbalances present in their source material (Bender et al. 2021). In biomedical contexts, these biases carry particular weight because they can directly influence clinical reasoning, treatment recommendations, and health outcomes. A recent systematic review of 24 studies evaluating demographic biases in medical LLMs found that 91.7% identified biases, with gender bias observed in 93.7% and racial or ethnic bias in 90.9% of studies examined (Omar et al. 2025). They noted that newer models similar to the ChatGPT family of models showed text-based demographic bias, for example when creating vignettes and discharge instructions. These models also showed bias in forecasting patient outcomes and to a lesser extent, diagnostic recommendations.

As such, clinical implications are tangible. Zack et al. (2024) demonstrated that GPT-4 does not appropriately model the demographic diversity of medical conditions, consistently producing clinical vignettes that stereotype demographic presentations. Differential diagnoses were more likely to include diagnoses that stereotype certain races, ethnicities, and genders, with significant associations found between demographic attributes and recommendations for more or fewer expensive procedures (Zack et al. 2024). Similarly, Bouguettaya et al. (2025) examined racial bias across four LLMs including Claude Sonnet 3.5, ChatGPT-4o, Gemini 1.5 Pro, and NewMes-v15 (Llama-3b variant) in psychiatric diagnosis and treatment, finding that LLMs often proposed inferior treatments when patient race was explicitly or implicitly indicated, though diagnostic decisions demonstrated minimal bias.

These findings have direct implications for the prompt engineering techniques discussed in this review. In Section 2.3, we demonstrated that persona assignment produces variable outputs that draw from composite archetypes within training data (Figure 4). When a prompt instructs the LLM to "take on the role of a clinician," the model draws from statistical patterns of clinician behavior in its training data, patterns that may encode implicit biases around race, gender, and socioeconomic status. The more specific the persona (e.g., "an emergency department physician in an urban hospital"), the more likely it may activate particular stereotypical associations.

Similarly, the few-shot examples discussed in Section 3 can inadvertently encode biases if they are drawn from a narrow demographic range. Our recommendation for example diversity in Figure 9, originally motivated by improving extraction robustness, also serves as a bias mitigation strategy: diverse, representative examples reduce the likelihood that the model will learn and reproduce demographic patterns from the prompt itself. This dynamic is well illustrated in the few-shot bias mitigation experiments reported by Sallami and Aïmeur (2025), who used a metric called Disparate Impact (DI) to measure whether unprivileged demographic groups (such as women, non-English speakers, or racial minorities) received equitable outcomes compared to privileged groups, with an acceptable DI score falling between 0.8 and 1.25, where scores below 0.8 indicate that the model is systematically disadvantaging certain groups. When GPT-4 was prompted with no examples at all (zero-shot), it produced a DI score of 0.92, meaning outputs still subtly disadvantaged certain demographic groups despite the absence of overt bias in the instruction. Crucially, as the authors introduced more few-shot examples into the prompt, the DI score rose progressively, reaching 1.10 with four examples. The composition of the examples supplied in the prompt, in other words, directly shaped whose experiences the model centred in its generations (Sallami & Aïmeur, 2025).

Beyond clinical bias, a significant concern for the global research community is the linguistic bias embedded in LLM training data. Training corpora are heavily skewed toward English-language content, and approximately 98% of scientific output is published in English. This creates a compounding disadvantage for researchers working in non-English languages and for biomedical knowledge from underrepresented regions. As we discussed in Section 2.4, writing scientific articles in English for non-native English speakers is difficult (Aydin et al. 2023) and while LLMs can help refine and proof-read scientific writing, improving accessibility (Smith et al. 2024), they also risk homogenizing scientific voice toward patterns that reflect English-dominant, Western-centric training data (Kobak et al. 2025).

### 9.2. Data Governance and Privacy

As researchers adopt LLMs for tasks ranging from literature summarization to data extraction, they routinely provide these models with potentially sensitive information, including unpublished research data, patient-derived datasets, and proprietary methodologies. The data governance implications of this practice are still emerging.

A well-documented cautionary example comes from outside academia: in 2023, Samsung engineers inadvertently leaked confidential source code and internal meeting notes through three separate incidents involving ChatGPT. The submitted data was retained by OpenAI for model training, effectively placing proprietary information into the hands of a third party (Culotta and Mattei 2024) . The same risk applies to researchers who paste patient data, clinical notes, or unpublished experimental results into commercial LLM web interfaces. Unlike local software tools, cloud-based LLMs may store, process, and potentially use submitted data for model improvement unless explicitly opted out with policies constantly being updated. Table 3 reflects the current data retention and usage policies for the 3 big models providers. By default, providers will train on user data even when the user has a paid subscription; however, opting out is an option. By default, providers do not train on API or enterprise data. Table 4 is a current snapshot as of late March 2026 and will most likely change in the future. We recommend the user look at the latest policies based on the model they wish to work with.

**Table 4. Default data retention and training policies for major LLM providers across subscription tiers (as of March 2026).** "Opt-in/out" indicates user data may be used for model training by default but can be disabled. "No" indicates data is not used for training. Retention periods refer to how long user-submitted prompts and responses are stored after deletion. Policies are subject to change.

| Provider | Tier | Data Retention (default) | Used for Training? |
|---|---|---|---|
| **Anthropic (Claude)** | Free / Pro / Max | 30 days post-deletion | Opt-in/out |
| | API | 30 days | No |
| | Enterprise | 30 days post-deletion; admin-controlled | No |
| **OpenAI (ChatGPT)** | Free / Plus / Pro | Indefinite while account active.<br>30 days after deletion. | Opt-in/out |
| | API | 30 days | No |
| | Enterprise | 30 days post-deletion; admin-controlled | No |
| **Google (Gemini)** | Free / Advanced | 18 months default | Opt-in/out |
| | API | 55 days | No |
| | Enterprise | Admin-controlled (3, 18, or 36 m) | No |

This concern is particularly acute in healthcare and biomedical research, where data protection is governed by regulations such as HIPAA in the United States and GDPR in Europe. A recent scoping review of 464 studies on LLM applications in healthcare using patient health information found that more than a third of studies failed to report any patient health information protection measures, that deidentification practices were frequently described in vague or non-transparent terms, and that reidentification risk assessment was almost entirely absent from the literature, suggesting that regulatory compliance on paper does not guarantee meaningful privacy protection in practice (X. Zhong et al. 2025). Jonnagaddala et al. (2025) outlined several privacy-preserving strategies for working with electronic health records in the LLM era, including locally deployed models, synthetic data generation, federated learning, differential privacy, and deidentification, recommending that the appropriate strategy be selected based on the specific LLM task, data type, and available computational infrastructure (Jonnagaddala and Wong 2025). Notably, in January 2025, the U.S. Department of Health and Human Services proposed the first major update to the HIPAA Security Rule in 20 years, introducing stricter expectations for risk management, encryption, and resilience specifically addressing AI systems that process protected health information.

These concerns connect directly to the prompt engineering practices discussed throughout this review. In Section 2.2, we advised that researchers may need to provide full research articles for effective processing, but this means sensitive content enters the LLM's context and potentially its training pipeline. The decomposition strategies presented in Section 6 involve sending multiple data fragments across separate queries, each of which may be stored independently. On the other hand, the agentic tools discussed in Section 7 such as Claude Code and Codex operate on local files and offer a fundamentally different privacy profile compared to web-based chat interfaces, though their data handling policies still require scrutiny.

From a practical standpoint, researchers should distinguish between API access (where data is typically not retained) and web chat interfaces (where data may be used for training by default). Institutions should develop clear policies regarding which data types may be processed through commercial LLMs, and when it comes to sensitive biomedical data, locally deployed open-source models or institutional AI infrastructure should be considered. A number of open-source models are available for download and deployment on either laptops, computers or more dedicated hardware (Table 5) that could be utilized by individuals and labs. This is not an exhaustive list but a snapshot of what a user can deploy on a work laptop or a workstation accessible to a lab.

**Table 5. Hardware requirements for running open-source LLMs locally across consumer and workstation platforms.** "Yes" and "No" indicate whether the model can be loaded and run on the given hardware configuration. Minimum memory refers to approximate RAM or VRAM required to load the model at default (FP16/BF16) precision; quantized variants may reduce requirements. Context refers to the maximum context window the model architecture supports, though effective context length may be constrained by available memory on smaller devices. Prices in GBP are accurate as of late March 2026.

| | | | **Platforms Comparison** | | | |
|---|---|---|---|---|---|---|
| **Model** | **Context*** | **Min Memory** | Lenovo Legion Pro 7i RTX 5090 · 24 GB £3,999 | MacBook Pro 14" M5 Pro · 24 GB £2,199 | Mac Studio M3 Ultra · 256 GB £6,199 | DGX Spark GB10 · 128 GB £4,200 |
| **Qwen 3.5-9B** | 262K | ~6 GB | Yes | Yes | Yes | Yes |
| **Nemotron 3 Nano** | 1M | ~18 GB | Yes | Yes | Yes | Yes |
| **Qwen 3.5-27B** | 262K | ~17 GB | No | No | Yes | Yes |
| **DeepSeek-R1-70B** | 128K | ~40 GB | No | No | Yes | Yes |
| **Nemotron 3 Super** | 1M | ~70 GB | No | No | Yes | Yes |
| **GPT-OSS-120B** | 128K | ~80 GB | No | No | Yes | Yes |

* Context refers to the maximum context window the model can support however in practice an M5 Pro running Qwen3.5-9B may support a context window of 8k before the token speed drops significantly.
** Prices are accurate as of 18th of March 2026

## 9.3. Responsible Usage and Journal Policies

The rapid adoption of LLMs in scientific research has prompted journals and publishing organizations to establish formal policies governing their use. As of early 2026, these policies span a wide spectrum, from outright prohibition to active encouragement, creating a fragmented landscape that researchers must navigate carefully. Table 6 summarizes the current AI policies of major journals and publishing bodies. A universal principle across all organizations is that AI tools cannot be credited as authors, since they cannot assume accountability for the accuracy, integrity, or originality of scholarly work. Beyond this consensus, policies diverge substantially.

**Table 6.** Comparison of AI usage policies across major journals and publishing organizations as of early 2026. Policies around Generative AI change often, refer to specific journal for up-to-date information.

| Organization | AI-Copy Editing (readability, polish, grammar, drafting) | AI as Author | Peer Review Use |
|---|---|---|---|
| ICMJE | Permitted with disclosure | Prohibited | Yes – must ensure confidentiality |
| COPE | Permitted with disclosure | Prohibited | Yes – disclose use |
| Nature | Permitted with disclosure | Prohibited | No |
| Science (AAAS) | Permitted with disclosure | Prohibited | No |
| NEJM AI | Encouraged | Prohibited | Yes – must ensure confidentiality |
| The Lancet | Permitted with disclosure | Prohibited | No |
| JAMA | Permitted with disclosure | Prohibited | Yes – must ensure confidentiality |

The ICMJE's 2026 revised recommendations represent the most comprehensive framework to date, introducing a dedicated section (Section V) addressing AI use across the full publishing workflow, from manuscript preparation to peer review and editorial decision-making. The guidelines require authors to disclose AI use in both the cover letter and the manuscript, verify all AI-generated content for accuracy, and confirm complete responsibility for the work. Critically, AI-generated material cannot serve as a primary source, and failure to disclose AI usage may constitute scientific misconduct ('ICMJE' 2026).

Yoo (2025), in a comparative review of editorial policies across major journals and organizations, highlighted a particularly important tension: AI detection tools often proposed as a solution for identifying undisclosed AI use frequently struggle with accuracy and may unfairly disadvantage non-native English writers. This creates an ironic situation in the context of our Section 2.4 discussion where the use of LLMs demonstrably helps non-native English speakers refine their scientific writing yet the resulting text may trigger AI detection flags precisely because it has been polished to a fluency level that detection algorithms associate with machine generation (J.-H. Yoo 2025).

The question of authorship in the age of AI raises deeper conceptual challenges. Akgun (2025) used the Ship of Theseus as a guiding metaphor: as human-authored components of a manuscript are progressively replaced with AI-generated content, at what point does the "author" cease to be the author? This idea has practical implications for how we think about the systematic prompt engineering approaches described throughout this review. The better the prompt, the more the LLM contributes to the final output, that is, the LLM can be more autonomous for longer, completing more before asking for user feedback. A well-engineered prompt that produces a comprehensive literature summary, or a polished draft may involve minimal post-hoc editing, certainly blurring the line between tool use and content generation (Akgün 2025).

As capabilities of Generative AI keep expanding and improving, so will the policies and the uncertainty around what is acceptable and when, will constantly change. Meanwhile, the prompt engineering community should develop best practices not only for generating high-quality outputs, but also for documenting and disclosing the role of AI in the research process. We recommend that researchers maintain prompt logs; records of the specific prompts, models, and parameters used during their work, as a form of methodological transparency analogous to recording reagent lot numbers or software versions. This may not be strictly necessary in the future where agents can traverse many tasks sequentially obtaining instructions from the context they observe, but at least a prompt log can serve as a way to help others in the immediate lab and community to understand and iterate on the technology.

## 10. Conclusion

This review was our attempt to synthesize and explore various high-impact prompt engineering ideas, with our observations and reflections while providing specific use cases relevant to academics within the life science. The Prompt Report describes 58 different prompt engineering techniques, here we focus on a small subset, providing rich examples, recommendations and references to other studies where prompt engineering techniques were developed and utilized. While it’s possible that next-generation of models may reduce the need for prompt engineering, so far the opposite has actually held true. Agentic systems like Claude Code, while highly capable, require explicit decomposition of tasks to stay on track. We hope that the case studies and examples explored in this review will help others to develop their own high-quality, field specific prompts that generate reliable, trustworthy and reproducible outputs for their field-specific research applications.

**Acknowledgements**
V.R. would like to acknowledge Prof. Emeritus Cristobal Dos Remedios for suggesting we publish the various ideas that we have presented here.

## Conflict of Interest

The authors have no relevant financial or non-financial interests to disclose.